%% file: iclr2027_conference.tex
\documentclass{article} 
\usepackage{iclr2027_conference,times}

\input{math_commands.tex}

\usepackage{hyperref}
\usepackage{url}
\usepackage{colortbl}
\usepackage{fvextra}
\DefineVerbatimEnvironment{ICRPrompt}{Verbatim}{
  fontsize=\footnotesize,
  breaklines=true,
  breakanywhere=true
}

\title{Beyond Final Accuracy:
Auditing Communication in LLM Multi-Agent Systems}

\author{
Shixuan Li$^{1}$\thanks{Equal contribution.}\ \ ,
Wei Yang$^{2}$\footnotemark[1]\ \ ,
Peiyu Zhang$^{1}$, Anzhe Cheng$^{1}$, Heng Ping$^{1}$, Paul Bogdan$^{1}$ \\
$^{1}$Ming Hsieh Department of Electrical and Computer Engineering \\
$^{2}$Thomas Lord Department of Computer Science \\
University of Southern California, Los Angeles, CA 90089, USA
}
\usepackage{xcolor}

\usepackage{booktabs}

\usepackage{graphicx}
\usepackage{wrapfig}
\usepackage{amsmath}
\usepackage{amssymb}
\usepackage{longtable}
\usepackage{caption}
\iclrfinalcopy
\begin{document}

\maketitle

\begin{abstract}
Multi-agent communication aims to help agents benefit
from one another's information.
Yet improvements in system performance leave a fundamental
ambiguity: do they reflect effective communication,
a favorable agent architecture, or simply additional reasoning?
Because communication methods are commonly evaluated within
the systems they were designed for, these factors are
difficult to disentangle.
Final accuracy further merges corrected errors and
corrupted answers into a single outcome, obscuring how
communication changes decisions.
We introduce Independent--Communicate--Revise (ICR),
a controlled framework that evaluates communication
as answer revision following independent reasoning.
ICR fixes initial reasoning trajectories, measures
correction and preservation conditional on both agents'
initial correctness, and uses a no-message revision
control to quantify gains beyond additional reasoning.
Across four reasoning benchmarks, our audit of textual
and latent communication reveals that similar aggregate
accuracy can conceal substantially different revision
behaviors.
Compared with transmitting answers alone, full reasoning
increases correction while reducing preservation on
all four benchmarks, so richer messages amplify
beneficial and harmful influence alike.
Receiver-policy comparisons on MedQA and GPQA-D further
show that a structured verification policy shifts every
channel toward greater preservation and lower correction,
while its effect on selectivity varies across channels
and tasks.
These findings challenge treating communication quality
as an intrinsic property of a channel.
ICR therefore recenters evaluation on selective revision,
providing a unified framework for examining how
message content and receiver policies jointly
produce benefits and harms. Code is available at \url{https://anonymous.4open.science/r/AgentICR-AA82/}.
\end{abstract}

\input{sections/introduction}
\input{sections/related_work}

\input{sections/methods}
\input{sections/experiments}

\section{Conclusion}
\label{sec:conclusion}

We introduced Independent--Communicate--Revise (ICR),
which audits communication through fixed initial
trajectories, correctness-conditioned revision metrics,
and a no-message control.
Across four benchmarks, correction and preservation
expose behavioral differences that aggregate accuracy
obscures: full reasoning trades preservation for
correction relative to answer-only messages, and
receiver policies shift this balance further.
Communication should therefore be evaluated by its
benefits and harms jointly, with conclusions tied to
the message, delivery interface, and receiver policy.


\newpage



\section*{Reproducibility statement}
Anonymous source code and supporting materials for reproducing the reported experiments are available at \url{https://anonymous.4open.science/r/AgentICR-AA82/}.




\bibliography{iclr2027_conference}
\bibliographystyle{iclr2027_conference}

\appendix
\input{sections/appendix}

\end{document}

%% file: math_commands.tex
\usepackage{amsmath,amsfonts,bm}

\def\eqref#1{equation~\ref{#1}}

\def\1{\bm{1}}

\DeclareMathAlphabet{\mathsfit}{\encodingdefault}{\sfdefault}{m}{sl}
\SetMathAlphabet{\mathsfit}{bold}{\encodingdefault}{\sfdefault}{bx}{n}



%% file: sections/introduction.tex
\section{Introduction}
\label{sec:introduction}

\begin{wrapfigure}{r}{0.4\textwidth}
    \centering
    \includegraphics[width=\linewidth]{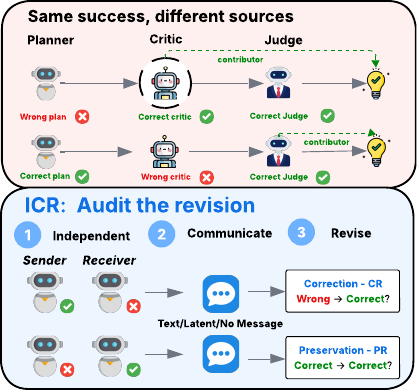}

    \caption{
Same final success, different intermediate contributions.
ICR audits correction and preservation.
    }
    \label{fig:intro_icr}
\end{wrapfigure}

Multi-agent systems extend large language models (LLMs)
through debate, role-based cooperation, and coordinated
workflows \citep{du2023improving,li2023camel,wu2023autogen,yang2025toward}.
Their communication increasingly reaches beyond explicit
text to hidden states and latent working memory
\citep{zou2025latent,peng2026statebridge,du2026enabling}.
This expansion of communication representations, however,
has outpaced our ability to evaluate their effects.
Communication mechanisms are commonly assessed together
with particular agent roles, interaction structures,
and inference budgets, making system-level improvements
difficult to attribute to information exchange.
The difficulty is most acute for latent communication,
where the exchanged information is not expressed in
natural language, yet evaluation must still establish
how it affects the receiver's decisions.
Even for textual messages, moreover, the same final
outcome can arise from different intermediate
contributions, which end-task accuracy alone does not
distinguish (Figure~\ref{fig:intro_icr}, top).

Our motivation comes from repeated executions of a
latent-communication pipeline, in which the same question
produced both correct and incorrect reasoning trajectories.
In our preliminary inspection, final answers frequently
followed choices made during early planning or critique,
and later stages often retained those choices.
Whether the final answer was correct thus depended less
on whether some agent had produced a useful solution
than on whether the others used it effectively.
Whereas sampling diverse reasoning paths can benefit
answer aggregation \citep{wang2022self}, communication
additionally requires deciding whether to revise an
existing answer.
Always following a peer permits both correction and
misdirection, while never changing rejects both harmful
interference and useful advice.
We therefore ask:
\emph{can communication correct an initially wrong answer
while preserving an initially correct one against
misleading peer information?}
This question applies to both textual and latent messages,
without requiring latent content to be decoded into text.

\setlength{\intextsep}{4pt}

We introduce \textbf{Independent--Communicate--Revise (ICR)},
a controlled framework for evaluating communication
as answer revision following independent reasoning
(Figure~\ref{fig:intro_icr}, bottom).
Agents first solve each question independently.
Their initial trajectories are then fixed and reused
across communication conditions, with each ordered
sender--receiver pair evaluated separately.
ICR measures the \emph{correction rate} (CR) of an incorrect
receiver paired with a correct sender and the
\emph{preservation rate} (PR) of a correct receiver
paired with an incorrect sender, while a matched
no-message revision control separates communication-driven
gains from those of additional reasoning alone
\citep{huang2024large}.
Correctness labels enter only offline selection,
stratification, and scoring, never message construction
or revision.
Because ICR standardizes the revision step while leaving
each channel's delivery mechanism intact and documented,
it gives textual and latent communication a shared
behavioral target and complements end-to-end evaluation
of complete systems.

We audit textual controls and representative latent
communication implementations across four reasoning
benchmarks.
Similar aggregate accuracy can conceal substantially
different correction--preservation profiles, in which
strong correction coexists with substantial corruption
and strong preservation with missed correction
opportunities.
These findings complement analyses of sycophancy and
premature agreement in multi-agent debate
\citep{yao2025peacemakertroublemakersycophancyshapes}
by conditioning evaluation on the initial correctness
of both participants.
Recovery when both agents are initially wrong also
remains rare in our experiments, which further
distinguishes correcting a receiver with the help of an
already correct peer from producing a correct answer
that neither agent had.

We further separate two dimensions of communication,
what is transmitted and how the receiver uses it.
Motivated by work on prompt compression
\citep{pan2024llmlingua2}, we compare No Message,
Answer Only, and Full Text, alongside a
content-independent random-delivery reference.
Across all four benchmarks, full reasoning increases
correction but reduces preservation relative to
transmitting answers alone.
Receiver-policy comparisons on MedQA and GPQA-D in turn
show that structured verification moves every channel
toward preservation at the expense of correction,
while the resulting selectivity changes vary across
channels and tasks.
Together, these experiments show that a channel's
revision behavior cannot be read apart from its message
content and receiver policy, and they provide a reusable
approach to evaluating selective revision.

Our contributions are threefold:
\begin{itemize}
    \item[(i)] We introduce ICR, a controlled framework
    that audits communication through fixed initial
    trajectories, correctness-conditioned metrics,
    and a no-message revision reference.

    \item[(ii)] We audit textual and latent communication
    on four reasoning benchmarks, exposing beneficial
    and harmful revision behaviors that aggregate
    accuracy conceals.

    \item[(iii)] We show how message content and receiver
    policy shape correction and preservation, using
    controlled comparisons and a random-delivery reference.
\end{itemize}

%% file: sections/related_work.tex
\section{Related Work}
\label{sec:related_work}

Our work connects three lines of research: multi-agent
collaboration, answer revision and reliability, and
communication representations, which together motivate
auditing how agents revise their answers under different
messages and interaction designs.

\textbf{Multi-agent collaboration and debate.}
LLM-based collaboration spans multi-agent debate
\citep{du2023improving,liang2024encouraging,xiong2023examining,yang2025learning}, role-playing
\citep{li2023camel}, programmable conversations
\citep{wu2023autogen}, and structured workflows
\citep{hong2024metagpt,chen2024agentverse,ping2026rem,ping2026verimoa,zhang2026oases,chen2026self,wang2025mixture,chen2026unitymas,yang2026tournament,li2026fortis}.
Communication topology also affects performance and
cost~\citep{zhang2025cut}, and sparse debate can match or
outperform fully connected interaction with fewer links
\citep{li2024improving}.
These studies show that interaction design shapes
system-level outcomes; ICR complements them by fixing
independently generated initial trajectories and comparing
communication conditions within a shared answer-revision
protocol.

\textbf{Answer revision and reliability.}
Reliable collaboration also requires deciding when an
existing answer should change.
Self-consistency aggregates sampled solutions
\citep{wang2022self}, Self-Refine uses iterative
self-feedback \citep{madaan2023self}, and Reflexion adds
feedback-driven reflection and episodic memory
\citep{shinn2023reflexion,ping2026poet,yang2026ramem}, yet intrinsic self-correction
can fail to improve reasoning and may degrade initially
correct responses \citep{huang2024large}.
In multi-agent settings, studies of sycophancy and identity
bias examine excessive agreement and preference for one's
own or a peer's answer
\citep{yao2025peacemakertroublemakersycophancyshapes,choi2026identity,yang2026auditing,yang2026adaptive};
Free-MAD counters conformity through consensus-free debate
and trajectory-based scoring \citep{cui2026free}, and
Minority Sentinel learns when to overturn majority voting
\citep{he2026minority}.
ICR builds on these concerns by jointly measuring correction
and preservation conditional on both agents' initial
correctness, with a matched no-message control that
separates communication-associated gains and losses from
those of additional reasoning alone.

\textbf{Communication representations and message content.}
LatentMAS enables latent reasoning and communication through
shared latent working memory \citep{zou2025latent}, and
StateBridge aligns sender hidden states with the receiver's
input space \citep{peng2026statebridge}, motivating evaluation
beyond textual messages.
For textual inputs, LLMLingua introduces budgeted prompt
compression \citep{jiang2023llmlingua}, LongLLMLingua
improves the presentation of relevant information in long
contexts \citep{jiang2024longllmlingua}, and LLMLingua-2
formulates extractive compression as token classification
\citep{pan2024llmlingua2}.
These studies ask what a message retains, but retained
content does not by itself support a correct revision.
ICR addresses this gap by comparing answer-only messages with
full sender reasoning and by separately varying the receiver's
revision policy, characterizing the benefits and harms of each
message, interface, and policy combination rather than ranking
textual and latent representations.

%% file: sections/methods.tex
\section{Independent--Communicate--Revise}
\label{sec:icr}

\begin{figure}[t]
    \centering
    \includegraphics[width=\linewidth]{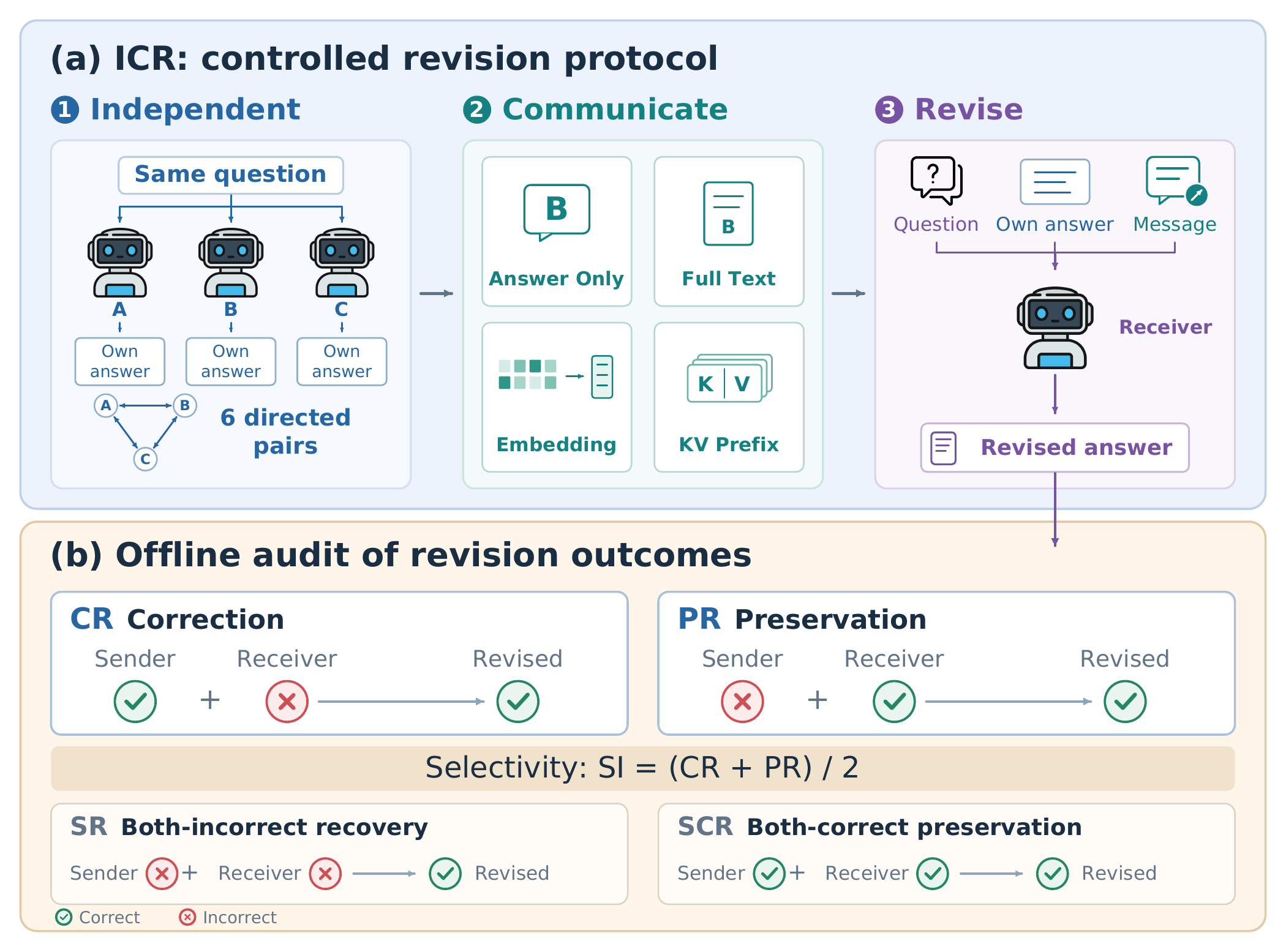}
    \caption{
    ICR overview.
    (a) Frozen initial trajectories support controlled
    communication and matched no-message revision.
    (b) Offline auditing separates four initial-correctness
    strata; each rate measures the fraction of correct
    revised answers within its stratum.
    }
    \label{fig:framework}
\end{figure}

The unit of evaluation in ICR is a directed revision event:
a receiver revisits its fixed initial solution using information
from a sender.
Figure~\ref{fig:framework} summarizes the controlled
revision protocol and the subsequent offline audit.

\subsection{Controlled Revision Protocol}
\label{sec:icr_protocol}
The protocol separates independent reasoning from
communication-driven revision, allowing communication
conditions to be compared from the same initial solutions.

\textbf{Independent generation and caching.}
For a question $q$ with reference answer $y$, let
$n$ agents independently produce initial records
\begin{equation}
    z_i = (r_i,a_i,h_i),
    \qquad i \in \{1,\ldots,n\},
\end{equation}
where $r_i$ is the reasoning trajectory, $a_i$ is the
extracted answer, and $h_i$ denotes associated
representations required by latent channels.
Independence means that agents do not observe one
another's outputs during initial generation;
it does not imply statistically independent errors.
These records are frozen and reused across conditions.

For each question, we construct all $n(n-1)$ ordered
sender--receiver pairs.
Our experiments use three agents, yielding six directions.
Each direction starts from the original cached records:
revised answers are never propagated into another
evaluation event.
Thus, the protocol evaluates single-hop revisions
without introducing sequential feedback.

\textbf{Communication and revision.}
For a direction $S \rightarrow R$, condition $c$
constructs a message from the sender's frozen record:
\begin{equation}
    m_{S \rightarrow R}^{(c)}
    = G_c(q,z_S).
    \label{eq:message_construction}
\end{equation}
The message can be textual or latent.
Let $D_c$ denote the channel-specific delivery operation,
such as text insertion, hidden-state injection,
or KV-prefix attachment.
The receiver's revised answer is
\begin{equation}
    a_R^{\prime(c,p)}
    =
    F_{\theta,p}
    \left(q,z_R;\,
    D_c\!\left(m_{S \rightarrow R}^{(c)}\right)\right),
    \label{eq:receiver_revision}
\end{equation}
where $\theta$ denotes the receiver model and $p$ its
revision policy.
The receiver retains access to the original question,
its initial reasoning, and its initial answer.

Channel comparisons hold the initial records, receiver
model, and revision policy fixed, while allowing message
construction and delivery to follow each channel.
Receiver-policy comparisons instead vary $p$ while
reusing messages and holding delivery fixed within
each channel.
Generation settings and seed configurations are recorded
for each comparison.
Reference answers and correctness labels are never
provided to message construction or revision.

\subsection{Auditing Revision Outcomes}
\label{sec:icr_metrics}

Let $\operatorname{Correct}(a,y)$ be the task-specific
correctness criterion.
Define initial and post-revision correctness as
\begin{equation}
    b_i = \mathbb{I}\!\left[
        \operatorname{Correct}(a_i,y)
    \right],
    \qquad
    b_R^{\prime(c,p)}
    = \mathbb{I}\!\left[
        \operatorname{Correct}(a_R^{\prime(c,p)},y)
    \right].
\end{equation}
Here, $\mathbb{I}[\cdot]$ is the indicator function,
equal to 1 when the enclosed statement is true and 0 otherwise.
We partition revision events by the receiver's and
sender's initial correctness, $(b_R,b_S)$.
These strata are determined before communication and
remain identical across compared conditions.
For brevity, we suppress $(c,p)$ below.

\textbf{Four initial correctness states.}
The two mixed-correctness strata measure whether
communication corrects errors and preserves correct answers:
\begin{equation}
\mathrm{CR}=\Pr(b'_R=1 \mid b_R=0,b_S=1),
\qquad
\mathrm{PR}=\Pr(b'_R=1 \mid b_R=1,b_S=0).
\label{eq:cr}
\end{equation}
CR is the \textbf{correction rate} of an initially wrong
receiver with a correct sender.
PR is the \textbf{preservation rate} of an initially
correct receiver with a wrong sender;
$1-\mathrm{PR}$ is its corruption rate.

The remaining strata measure
\textbf{both-incorrect recovery} and
\textbf{both-correct preservation}:
\begin{equation}
\mathrm{SR}=\Pr(b'_R=1 \mid b_R=0,b_S=0),
\qquad
\mathrm{SCR}=\Pr(b'_R=1 \mid b_R=1,b_S=1).
\label{eq:sr_scr}
\end{equation}
SR distinguishes recovery without an initially correct
participant from correction using an already correct peer.
SCR measures whether revision preserves correctness
when both participants start correct.
Each rate is estimated as the fraction of correct
post-revision outcomes within its stratum;
empty strata are reported as unavailable.
Sender correctness refers to its initial answer,
not the validity of every claim in its message.

\textbf{Selectivity and its reference points.}
We summarize the mixed-correctness strata with the
equal-weight \textbf{selectivity index}:
\begin{equation}
    \mathrm{SI}
    = \frac{\mathrm{CR}+\mathrm{PR}}{2},
    \qquad
    \mathrm{SI}-\frac{1}{2}
    = \frac{\mathrm{CR}-(1-\mathrm{PR})}{2}.
    \label{eq:si}
\end{equation}
SI therefore balances correction against corruption
with equal weight assigned to the two strata.
Always retaining the initial answer gives
$(\mathrm{CR},\mathrm{PR},\mathrm{SI})=(0,1,0.5)$.
Always copying the sender's answer on mixed-correctness
pairs gives $(1,0,0.5)$.
These analytical references illustrate why SI must be
reported together with CR and PR:
the same summary can describe very different behaviors.
SI characterizes observed revision outcomes under a
specified channel and receiver policy; it does not
directly measure an internal ability to recognize truth.

\textbf{Relation to aggregate accuracy.}
Evaluating both directions of every mixed-correctness
pair gives equal CR and PR denominators.
Accuracy on these mixed directions consequently equals SI.
For a broader evaluation population, let
$w_{rs}=\Pr(b_R=r,b_S=s)$ denote its initial-state proportions.
Post-revision accuracy decomposes as
\begin{equation}
    \mathrm{Acc}_{\mathrm{rev}}
    =
    w_{01}\mathrm{CR}
    + w_{10}\mathrm{PR}
    + w_{00}\mathrm{SR}
    + w_{11}\mathrm{SCR}.
    \label{eq:accuracy_decomposition}
\end{equation}
Aggregate accuracy thus depends on both revision behavior
and the prevalence of each initial state.
All weights and rates in this identity must refer to
the same evaluation population.
We distinguish measured subset accuracy from estimated
full-set accuracy when some strata are only sampled.

\subsection{Comparison Conditions and Attribution Controls}
\label{sec:icr_controls}
We compare alternative messages and delivery interfaces
against a no-message revision control to assess what
communication adds beyond additional reasoning.

\textbf{Communication conditions.}
Our instantiation includes Answer Only, which supplies
the sender's normalized final answer, and Full Text,
which supplies its reasoning and answer.
Latent conditions adapt aligned hidden-state transfer
\citep{peng2026statebridge} and sender-derived latent
working memory \citep{zou2025latent}
to the same directed revision setting.
These are adaptations of communication mechanisms,
rather than reproductions of their full native systems.
Their representations, injection locations, and
additional computation are documented in
Appendix~\ref{app:channels}.

\textbf{No-message and paired comparisons.}
No Message retains the question, receiver prior,
and revision opportunity but supplies no peer information.
It differs from Keep Initial, which performs no revision.
For every evaluated event, we compare a communication
outcome with its matched no-message outcome under
the same receiver policy.

On a common set of $N$ events, let $N_{\mathrm{gain}}$
count outcomes that are correct with communication
but incorrect without it, and let $N_{\mathrm{loss}}$
count the reverse.
The paired accuracy difference is
\begin{equation}
    \Delta\mathrm{Acc}
    =
    \frac{N_{\mathrm{gain}}-N_{\mathrm{loss}}}{N}.
    \label{eq:paired_gain_loss}
\end{equation}
We apply this decomposition both to the evaluated
population and to individual correctness strata.
It identifies the gains and losses relative to
additional reasoning without a message.
Policy comparisons use a separate no-message reference
for each policy.

\textbf{Uncertainty and scope.}
We bootstrap questions, keeping their associated
directions and compared conditions together.
This preserves within-question dependence and pairing;
the resulting intervals do not capture variation across
unobserved generation seeds.
ICR standardizes the revision experiment and audit
targets while retaining channel-specific interfaces
and computation.
Its conclusions therefore concern the evaluated
channel--policy combinations, rather than an intrinsic
ranking of communication representations.

%% file: sections/experiments.tex
\section{Experiments}
\label{sec:experiments}

\input{tables/main_results}
We organize the experiments around three questions:
what final accuracy conceals, how message content affects
revision, and how receiver policy shapes correction
and preservation.

\subsection{Experimental Setup}
\label{sec:experimental_setup}

\textbf{Tasks and evaluation population.}
We evaluate Qwen3-4B~\citep{yang2025qwen3} on MedQA~\citep{jin2021disease}, ARC-Challenge~\citep{clark2018think}, GSM8K~\citep{cobbe2021training},
and GPQA-Diamond (GPQA-D)~\citep{rein2023gpqa}, with HumanEval+~\citep{liu2023your} in the appendix.
Retaining questions with at least one initially incorrect
agent yields 430 questions and 2,580 directed events,
including 840 mixed-correctness directions.
Additional Qwen3-8B experiments on MedQA and GPQA-D
use model-specific initial trajectories and contain
46 and 54 mixed-correctness questions, yielding
92 and 108 events per CR/PR stratum, respectively.

\textbf{Comparison conditions.}
The main audit evaluates \textbf{No Message},
\textbf{Answer Only}, \textbf{Full Text},
\textbf{StateBridge}, and \textbf{LatentMAS}
under Critical Evaluation.
Initial trajectories and receiver instructions are fixed
across conditions within each model.
For Qwen3-4B, Answer Only covers mixed-correctness directions,
while the other conditions cover all retained directions.
For Qwen3-8B, all five conditions cover mixed-correctness
directions only.
Delivery interfaces are detailed in
Appendix~\ref{app:channels};
receiver-policy comparisons appear in
Section~\ref{sec:receiver_policy}.

\textbf{Metrics and statistical analysis.}
We report CR, PR, and SI, with SR and SCR for the
Qwen3-4B conditions covering all retained directions.
Estimated full-set accuracy (Acc.$^{\dagger}$) uses
assumptions for unobserved outcomes. To reduce cost, Qwen3-4B revisions on all-three-correct questions are partially evaluated, with unobserved outcomes extrapolated or assumed correct
(Appendix~\ref{app:extrapolation}). 
Intervals use question-cluster bootstrap resampling,
jointly retaining associated directions and compared
conditions.
Evaluation details are provided in
Appendix~\ref{app:scoring}.

\subsection{RQ1: What Does Accuracy Conceal?}
\label{sec:communication_audit}

\input{tables/joint_states}

\textbf{Correction and preservation reveal distinct profiles.}
For Qwen3-4B, Table~\ref{tab:main_results}
separates correction from preservation.
On MedQA, Full Text achieves a CR of 80.17\% but a PR
of 42.24\%, whereas StateBridge achieves a lower CR
of 54.31\% and a higher PR of 71.55\%.
These profiles yield SI values of 61.21\% and 62.93\%,
respectively.
The relative performance also varies across tasks:
Full Text has the highest observed SI on ARC-C and GPQA-D,
while StateBridge has the highest on MedQA and GSM8K.
No Message exhibits the highest PR among the evaluated
revision conditions, but substantially lower CR.

\begin{figure}[t]
\centering
\includegraphics[width=0.92\linewidth]{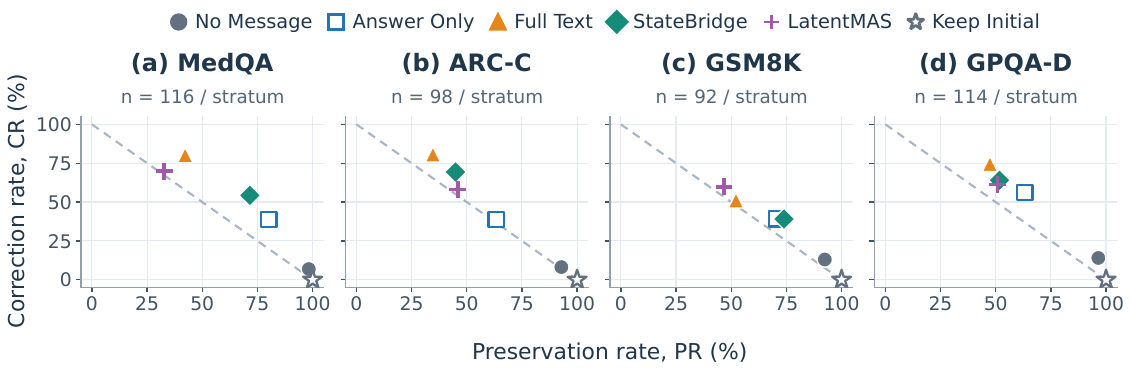}
\caption{
Qwen3-4B Correction--preservation profiles.
Markers show point estimates; the dashed line indicates
$\mathrm{CR}+\mathrm{PR}=100\%$, corresponding to
$\mathrm{SI}=50\%$.
The receiver prompt is fixed, while message content
and delivery vary.
}
\label{fig:cr_pr}
\vspace{-9pt}
\end{figure}

\textbf{Aggregate accuracy can obscure these differences.}
On GSM8K, estimated full-set accuracy ranges from 94.68\%
to 94.79\% across the evaluated revision conditions,
a spread of only 0.11 percentage points.
Among the communicating conditions, however,
PR ranges from 46.74\% to 73.91\%, a gap of 27.17 points.
Figure~\ref{fig:cr_pr} visualizes the correction--preservation
profiles that aggregate accuracy does not expose.
Because full-set accuracy includes a large population
of all-correct questions, it need not closely track
performance on mixed-correctness pairs.

\textbf{Paired gains and losses.}
On MedQA, Full Text produces 86 correct outcomes where
No Message is wrong, but 75 incorrect outcomes where
No Message is correct, giving a net gain of 11 out of
720 retained revision events.
StateBridge has 56 correct outcomes where No Message
is wrong and 41 incorrect outcomes where No Message
is correct, giving an observed net gain of 15
out of 720 retained events.
On GSM8K, the corresponding net changes are
$-5$ for Full Text, $+4$ for StateBridge, and $-4$
for LatentMAS out of 564 events.
Thus, a higher CR does not necessarily translate
into higher retained-subset accuracy:
correction gains can be offset by losses in other
initial correctness states.
Appendix~\ref{app:paired} reports the complete paired gain/loss results
over retained directions.

\textbf{Recovery remains limited when both agents
are initially wrong.}
Across the four main datasets, SR for the communicating
conditions ranges from 0.30\% to 2.55\%
(Table~\ref{tab:joint_states}).
It is numerically below No Message in every comparison
except StateBridge on GPQA-D, which recovers 11 cases
versus 10 for No Message.
Within the retained population, SCR remains high:
Full Text preserves correctness in every evaluated
both-correct pair.
Together, these results distinguish recovering a correct
answer when neither agent initially has one from correcting
a receiver whose sender is already correct.
Detailed answer transitions are provided in
Appendix~\ref{app:transitions}.

\textbf{Additional model scale.}
The Qwen3-8B results in Table~\ref{tab:main_results}
show the same directional contrast between Full Text
and Answer Only: higher CR and lower PR on both tasks.
StateBridge attains the highest observed SI on MedQA,
where its paired SI gain over No Message has a positive
95\% interval.
On GPQA-D, Answer Only has the highest SI point estimate.
Communication profiles thus remain task-dependent at
the larger model scale.
Detailed results are reported in
Appendix~\ref{app:qwen3_8b}.

\subsection{RQ2: How Does Message Content Affect Revision?}
\label{sec:message_contribution}

\begin{wraptable}{r}{0.53\textwidth}
\centering
\caption{
Full Text minus Answer Only, in percentage points.
95\% CIs for $\Delta$SI use 10,000 paired
question-cluster bootstrap resamples.
}
\small
\setlength{\tabcolsep}{3pt}
\renewcommand{\arraystretch}{1.05}
\captionsetup{font=footnotesize,skip=4pt}

\begin{tabular}{@{}lrrrr@{}}
\toprule
Dataset
& $\Delta$CR
& $\Delta$PR
& $\Delta$SI
& 95\% CI \\
\midrule
MedQA
& $+41.38$ & $-37.93$ & $+1.72$
& $[-4.31,\;7.76]$ \\
ARC-C
& $+41.84$ & $-28.57$ & $+6.63$
& $[-0.51,\;13.27]$ \\
GSM8K
& $+11.96$ & $-18.48$ & $-3.26$
& $[-12.50,\;6.52]$ \\
GPQA-D
& $+18.42$ & $-15.79$ & $+1.32$
& $[-6.14,\;8.77]$ \\
\bottomrule
\end{tabular}

\label{tab:message_effects}
\end{wraptable}

For Qwen3-4B, we compare No Message, Answer Only, and Full Text on
the same mixed-correctness directed pairs, with initial
trajectories and the receiver revision policy fixed.
This comparison examines answer exposure and the additional
effect of supplying the sender's full reasoning,
including its accompanying increase in message length.

\textbf{More correction comes with less preservation.}
Relative to Answer Only, Full Text increases CR by
11.96--41.84 percentage points, while decreasing PR
by 15.79--37.93 points
(Table~\ref{tab:message_effects}).
The paired 95\% bootstrap intervals for CR increases
exclude zero on MedQA, ARC-C, and GPQA-D; on GSM8K, the interval includes zero.
The intervals for PR decreases exclude zero on all
four datasets.
Detailed CR and PR difference intervals are reported in
Appendix~\ref{app:message_metric_differences}.
Full reasoning therefore shifts the balance toward more correction and less preservation.




\begin{figure}[t]
\centering
\includegraphics[width=0.93\linewidth]
{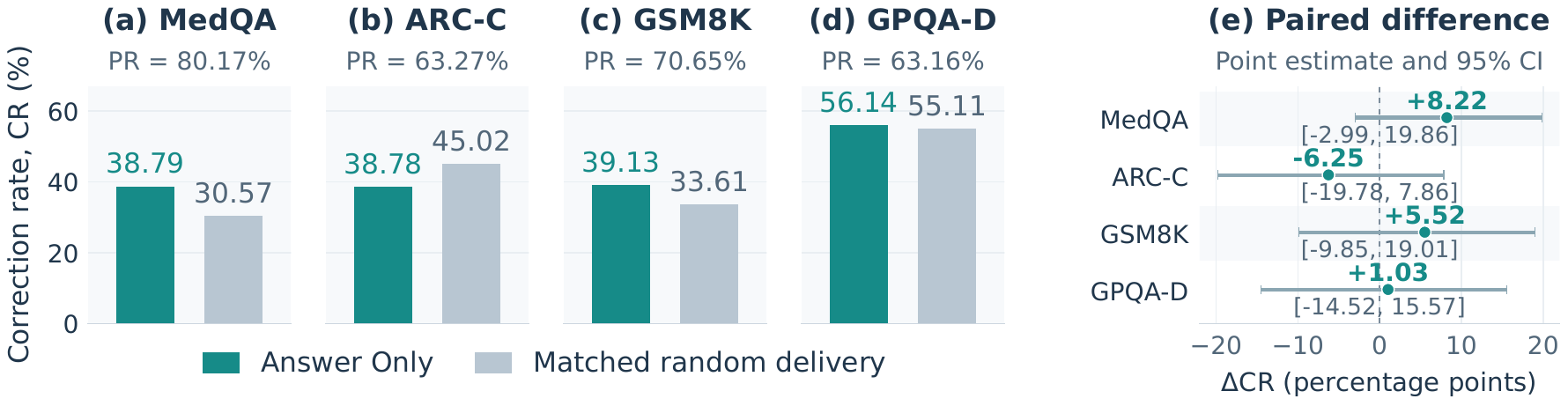}
\caption{
Qwen3-4B Answer Only versus random delivery at matched PR.
(a--d) CR for Answer Only and a reference mixing Full Text
and No Message to match its PR.
(e) Answer Only minus reference CR, with 95\% paired
question-cluster bootstrap intervals.
}
\label{fig:message_contribution}
\end{figure}


\textbf{Answer matching increases in both correctness strata.}
On MedQA, Full Text increases sender-matching outputs
from 45 to 93 in the CR stratum and from 23 to 67 in
the PR stratum relative to Answer Only.
Valid-output answer-change rates also increase across
all four datasets.
These patterns indicate more frequent revision, while
sender matching alone does not establish causal adoption.

\textbf{Comparison with random message delivery.}
We next ask whether the Answer Only profile exceeds
a content-independent reduction in message exposure.
Consider a policy that uses Full Text with probability
$\alpha$ and No Message otherwise, independently of
the question and correctness labels.
Its expected rates are
\begin{equation}
\begin{aligned}
\mathrm{CR}_{\alpha}
&=\alpha\,\mathrm{CR}_{\mathrm{Full}}
 +(1-\alpha)\,\mathrm{CR}_{\mathrm{None}},
\qquad
\mathrm{PR}_{\alpha}
=\alpha\,\mathrm{PR}_{\mathrm{Full}}
 +(1-\alpha)\,\mathrm{PR}_{\mathrm{None}}.
\end{aligned}
\end{equation}
For each dataset, we match this reference to the PR
of Answer Only and compare the corresponding CR
(Figure~\ref{fig:message_contribution}).
Detailed reference values are reported in
Appendix~\ref{app:random_delivery}.
All four intervals include zero, leaving the CR differences
at matched preservation unresolved.

\subsection{RQ3: How Does Receiver Policy Affect Revision?}
\label{sec:receiver_policy}

We compare two receiver policies to examine how revision
instructions shape the use of peer information in Qwen3-4B.
\textbf{Critical Evaluation}, used in the main audit,
asks the receiver to critically assess its previous
reasoning and the external message.
\textbf{Structured Verification} specifies an ordered
procedure: assess external claims against the problem,
check the receiver's own reasoning, and decide using
the surviving claims.

The comparison uses the same initial trajectories,
sender messages, and delivery interfaces within each
channel.
We evaluate mixed-correctness directions on MedQA
and GPQA-D, with 116 and 114 events per correctness
stratum, respectively.
We report changes in CR, PR, and SI, together with
communication increments relative to each policy's
No Message control.

\begin{table}[t]
\centering
\small
\caption{
Receiver-policy comparison on mixed-correctness directions with Qwen3-4B.
CR, PR, and SI are percentages.
Policy differences are computed as Structured Verification
minus Critical Evaluation and reported in percentage points.
The 95\% intervals for $\Delta$SI use 10,000 paired
question-cluster bootstrap resamples.
}

\setlength{\tabcolsep}{3pt}
\renewcommand{\arraystretch}{1.12}
{%
\begin{tabular}{@{}llrrrrrrrl@{}}
\toprule
& & \multicolumn{3}{c}{Critical Evaluation}
& \multicolumn{3}{c}{Structured Verification}
& \multicolumn{2}{c}{Policy difference} \\
\cmidrule(lr){3-5}
\cmidrule(lr){6-8}
\cmidrule(lr){9-10}
Dataset & Channel
& CR & PR & SI
& CR & PR & SI
& $\Delta$SI & 95\% CI \\
\midrule

MedQA
& No Message
& 6.90 & 98.28 & 52.59
& 3.45 & 99.14 & 51.29
& $-1.29$ & $[-3.88,\;1.29]$ \\

& Full Text
& 80.17 & 42.24 & 61.21
& 56.90 & 67.24 & 62.07
& $+0.86$ & $[-5.17,\;6.90]$ \\

& StateBridge
& 54.31 & 71.55 & 62.93
& 27.59 & 85.34 & 56.47
& $-6.47$ & $[-11.64,\;-1.29]$ \\

& LatentMAS
& 69.83 & 32.76 & 51.29
& 43.97 & 78.45 & 61.21
& $+9.91$ & $[1.29,\;18.53]$ \\

\midrule

GPQA-D
& No Message
& 14.04 & 96.49 & 55.26
& 7.02 & 97.37 & 52.19
& $-3.07$ & $[-7.02,\;0.88]$ \\

& Full Text
& 74.56 & 47.37 & 60.96
& 55.26 & 61.40 & 58.33
& $-2.63$ & $[-9.21,\;4.39]$ \\

& StateBridge
& 64.04 & 51.75 & 57.89
& 50.88 & 70.18 & 60.53
& $+2.63$ & $[-4.39,\;9.21]$ \\
& LatentMAS
& 61.40 & 50.88 & 56.14
& 43.86 & 74.56 & 59.21
& $+3.07$ & $[-4.39,\;10.53]$ \\
\bottomrule
\end{tabular}%
}

\label{tab:receiver_policy}
\end{table}

\textbf{Correction and preservation across receiver policies.}
Across both datasets, Structured Verification yields lower CR
and higher PR than Critical Evaluation for all three
communication conditions (Table~\ref{tab:receiver_policy}).
The resulting SI changes depend on the channel and task:
on MedQA, SI decreases for StateBridge and increases for
LatentMAS, with both intervals excluding zero;
on GPQA-D, all three SI difference intervals include zero.
Thus, greater preservation is accompanied by reduced correction.

\textbf{Communication increments relative to No Message.}
We measure communication increments relative to each policy's
no-message baseline as
$G(c,p)=\mathrm{SI}(c,p)-\mathrm{SI}(\text{No Message},p)$.
On MedQA, moving from Critical Evaluation to Structured
Verification increases $G$ for LatentMAS, with the 95\% interval
excluding zero.
Thus, its improvement extends beyond the change in
no-message revision performance. Complete communication-increment results are reported in
Appendix~\ref{app:policy_communication_increments}.

\textbf{Relative channel performance across policies.}
Channel comparisons should specify the receiver policy
and assess correction and preservation jointly,
using a policy-specific No Message reference.
Detailed SI contrasts between StateBridge and Full Text
are reported in Appendix~\ref{app:channel_policy_comparison}.

%% file: tables/main_results.tex
\begin{table}[t]
\centering
\caption{
Communication audit under Critical Evaluation (\%).
Acc.$^{\dagger}$ denotes full-set accuracy estimated
under assumptions about unobserved revision outcomes;
Initial answers reports measured accuracy.
Estimation assumptions are detailed in
Appendix~\ref{app:extrapolation}.
Bold marks the highest value per metric and dataset
among revision conditions within each model.
}
\setlength{\tabcolsep}{3pt}
\renewcommand{\arraystretch}{1.16}

\resizebox{\linewidth}{!}{%
\begin{tabular}{@{}lrrrrrrrrrrrrrrrr@{}}
\toprule
\multicolumn{17}{@{}l}{\textbf{(a) Qwen3-4B}} \\
\midrule

\rowcolor{blue!7}
& \multicolumn{4}{c}{\textbf{MedQA}}
& \multicolumn{4}{c}{\textbf{ARC-C}}
& \multicolumn{4}{c}{\textbf{GSM8K}}
& \multicolumn{4}{c}{\textbf{GPQA-D}} \\
\cmidrule(lr){2-5}
\cmidrule(lr){6-9}
\cmidrule(lr){10-13}
\cmidrule(lr){14-17}
Condition
& Acc.$^{\dagger}$ & CR & PR & \textcolor{blue!60!black}{SI}
& Acc.$^{\dagger}$ & CR & PR & \textcolor{blue!60!black}{SI}
& Acc.$^{\dagger}$ & CR & PR & \textcolor{blue!60!black}{SI}
& Acc.$^{\dagger}$ & CR & PR & \textcolor{blue!60!black}{SI} \\
\midrule

\rowcolor{black!4}
Keep Initial
& -- & 0.00 & 100.00 & 50.00
& -- & 0.00 & 100.00 & 50.00
& -- & 0.00 & 100.00 & 50.00
& -- & 0.00 & 100.00 & 50.00 \\

\rowcolor{black!4}
Initial answers
& 69.33 & -- & -- & --
& 93.91 & -- & -- & --
& 94.62 & -- & -- & --
& 54.04 & -- & -- & -- \\
\midrule

No Message
& 70.28 & 6.90 & \textbf{98.28} & 52.59
& 93.93 & 8.16 & \textbf{92.86} & 50.51
& 94.74 & 13.04 & \textbf{92.39} & 52.72
& 54.58 & 14.04 & \textbf{96.49} & 55.26 \\

Answer Only
& -- & 38.79 & 80.17 & 59.48
& -- & 38.78 & 63.27 & 51.02
& -- & 39.13 & 70.65 & 54.89
& -- & 56.14 & 63.16 & 59.65 \\

Full Text
& 70.89 & \textbf{80.17} & 42.24 & 61.21
& \textbf{94.13} & \textbf{80.61} & 34.69 & \textbf{57.65}
& 94.68 & 51.09 & 52.17 & 51.63
& \textbf{56.32} & \textbf{74.56} & 47.37 & \textbf{60.96} \\

StateBridge
& \textbf{71.11} & 54.31 & 71.55 & \textbf{62.93}
& 94.11 & 69.39 & 44.90 & 57.14
& \textbf{94.79} & 39.13 & 73.91 & \textbf{56.52}
& 55.75 & 64.04 & 51.75 & 57.89 \\

LatentMAS
& 69.89 & 69.83 & 32.76 & 51.29
& 93.52 & 58.16 & 45.92 & 52.04
& 94.69 & \textbf{59.78} & 46.74 & 53.26
& 55.06 & 61.40 & 50.88 & 56.14 \\

\midrule
\multicolumn{17}{@{}l}{\textbf{(b) Qwen3-8B}} \\
\midrule

\rowcolor{blue!7}
& \multicolumn{8}{c}{\textbf{MedQA}}
& \multicolumn{8}{c}{\textbf{GPQA-D}} \\[-2pt]
\cmidrule(lr){2-9}
\cmidrule(lr){10-17}
Condition
& \multicolumn{2}{r}{Acc.$^{\dagger}$}
& \multicolumn{2}{r}{CR}
& \multicolumn{2}{r}{PR}
& \multicolumn{2}{r}{\textcolor{blue!60!black}{SI}}
& \multicolumn{2}{r}{Acc.$^{\dagger}$}
& \multicolumn{2}{r}{CR}
& \multicolumn{2}{r}{PR}
& \multicolumn{2}{r}{\textcolor{blue!60!black}{SI}} \\[-2pt]
\midrule

\rowcolor{black!4}
Initial answers
& \multicolumn{2}{r}{78.89}
& \multicolumn{2}{r}{--}
& \multicolumn{2}{r}{--}
& \multicolumn{2}{r}{--}
& \multicolumn{2}{r}{57.91}
& \multicolumn{2}{r}{--}
& \multicolumn{2}{r}{--}
& \multicolumn{2}{r}{--} \\[-2pt]
\midrule

No Message
& \multicolumn{2}{r}{78.94}
& \multicolumn{2}{r}{7.61}
& \multicolumn{2}{r}{\textbf{93.48}}
& \multicolumn{2}{r}{50.54}
& \multicolumn{2}{r}{58.59}
& \multicolumn{2}{r}{14.81}
& \multicolumn{2}{r}{\textbf{92.59}}
& \multicolumn{2}{r}{53.70} \\[-2pt]

Answer Only
& \multicolumn{2}{r}{79.78}
& \multicolumn{2}{r}{31.52}
& \multicolumn{2}{r}{85.87}
& \multicolumn{2}{r}{58.70}
& \multicolumn{2}{r}{\textbf{59.43}}
& \multicolumn{2}{r}{50.93}
& \multicolumn{2}{r}{65.74}
& \multicolumn{2}{r}{\textbf{58.33}} \\[-2pt]

Full Text
& \multicolumn{2}{r}{80.11}
& \multicolumn{2}{r}{\textbf{77.17}}
& \multicolumn{2}{r}{46.74}
& \multicolumn{2}{r}{61.96}
& \multicolumn{2}{r}{59.26}
& \multicolumn{2}{r}{\textbf{70.37}}
& \multicolumn{2}{r}{44.44}
& \multicolumn{2}{r}{57.41} \\[-2pt]

StateBridge
& \multicolumn{2}{r}{\textbf{80.78}}
& \multicolumn{2}{r}{60.87}
& \multicolumn{2}{r}{76.09}
& \multicolumn{2}{r}{\textbf{68.48}}
& \multicolumn{2}{r}{59.01}
& \multicolumn{2}{r}{65.74}
& \multicolumn{2}{r}{46.30}
& \multicolumn{2}{r}{56.02} \\[-2pt]

LatentMAS
& \multicolumn{2}{r}{79.67}
& \multicolumn{2}{r}{73.91}
& \multicolumn{2}{r}{41.30}
& \multicolumn{2}{r}{57.61}
& \multicolumn{2}{r}{59.01}
& \multicolumn{2}{r}{44.44}
& \multicolumn{2}{r}{67.59}
& \multicolumn{2}{r}{56.02} \\

\bottomrule
\end{tabular}%
}

\label{tab:main_results}
\end{table}

%% file: tables/joint_states.tex
\begin{wraptable}[23]{r}{0.60\textwidth}
\centering
\caption{
Recovery and preservation by initial sender--receiver
correctness (\%).
Rates are measured on retained questions with at least
one initially incorrect agent among the three.
Answer Only is evaluated only on mixed-correctness
pairs and is omitted here.
}
\small

\setlength{\tabcolsep}{3pt}
\renewcommand{\arraystretch}{1.05}
\captionsetup{font=small,skip=4pt}

\resizebox{0.92\linewidth}{!}{%
\begin{tabular}{@{}lrrrr@{}}
\toprule
\multicolumn{5}{@{}l}{
\textbf{(a) Both-incorrect recovery (SR)}
} \\
\midrule
Condition & MedQA & ARC-C & GSM8K & GPQA-D \\
\midrule
No Message
& 2.52 & 1.52 & 1.80 & 2.31 \\
Full Text
& 0.46 & 0.30 & 0.60 & 1.39 \\
StateBridge
& 0.46 & 0.30 & 0.90 & 2.55 \\
LatentMAS
& 1.61 & 0.91 & 0.60 & 0.93 \\

\midrule
\multicolumn{5}{@{}l}{
\textbf{(b) Both-correct preservation (SCR)}
} \\
\midrule
Condition & MedQA & ARC-C & GSM8K & GPQA-D \\
\midrule
No Message
& 100.00 & 93.75 & 97.83 & 91.67 \\
Full Text
& 100.00 & 100.00 & 100.00 & 100.00 \\
StateBridge
& 100.00 & 98.44 & 97.83 & 97.92 \\
LatentMAS
& 100.00 & 95.31 & 95.65 & 95.83 \\
\bottomrule
\end{tabular}%
}

\label{tab:joint_states}
\end{wraptable}

%% file: sections/appendix.tex
\newpage
\section{Implementation and Reproducibility}

\subsection{Model and Generation Settings}

The primary experiments use \texttt{Qwen/Qwen3-4B}.
Additional experiments with \texttt{Qwen/Qwen3-8B}
are described in Appendix~\ref{app:qwen3_8b}.
Inference uses \texttt{bfloat16}, caching, and one sequence
per worker process.
The software environment comprises Python 3.12.3,
Transformers 4.51.3, and PyTorch 2.7.1 with CUDA 12.8.
The main-run manifests record four NVIDIA RTX 5090 GPUs
with 32\,GB memory each.
The attention implementation is selected by the library default.

\paragraph{Decoding.}
We use sampling with temperature $0.6$, top-$p=0.95$,
and top-$k=20$.
The top-$k$ value is inherited from the model's generation
configuration.
Generation stops at a configured end-of-sequence token
or the token limit.
Both stages use the model's chat template with thinking enabled.
The fixed system message is:

\begin{ICRPrompt}
You are Qwen, created by Alibaba Cloud. You are a helpful assistant.
\end{ICRPrompt}

\paragraph{Response processing.}
The rendered prompt opens the assistant response with
\texttt{<think>}.
The thinking segment is removed before answer parsing
and textual reuse.
The receiver's prior reasoning is therefore its stored
post-thinking response, and Full Text transmits the sender's
post-thinking response verbatim, without further shortening.
Latent payload construction is described in Appendix~B.

\paragraph{Generation budgets.}
The final configuration permits 16,384 new tokens in both
stages.
During data collection, lower limits of 2,048, 4,096, or
8,192 tokens were increased as needed.
Responses that had already terminated were retained.
All retained initial responses generated under lower limits
had terminated before reaching those limits.

Eight revision records remained truncated at an earlier limit:
four on ARC-C, one on GSM8K, and three on HumanEval+.
Two belong to the ARC-C StateBridge preservation stratum;
the remaining six belong to the both-incorrect stratum.
These records are retained and evaluated using the same
answer-parsing and scoring rules as other records.
A non-termination flag does not itself determine correctness;
an output without a parseable answer is scored incorrect.

\paragraph{Randomness.}
Before each generation, the implementation initializes
the Python, NumPy, PyTorch, and CUDA random generators
through \texttt{transformers.set\_seed}.
Per-record seeds are derived deterministically from the
global seed 42, a run identifier, the question identifier,
and either the agent identifier or the sender--receiver direction.
The derivation excludes the communication condition
and receiver-policy label.
Deterministic GPU algorithms are not enforced.

\subsection{Independent Generation Prompts}

The independent stage uses task-specific user prompts.
Below, braces denote fields populated from the benchmark.
Boxed-answer syntax is shown as rendered for the model.
Line wrapping is adjusted for presentation.

\paragraph{MedQA.}
\begin{ICRPrompt}
You are an independent problem-solving agent.

Solve the medical multiple-choice question carefully and independently.

Reason from the evidence in the question.
Do not assume another agent will review your answer.

At the end, return exactly one final option in benchmark-compatible form: \boxed{A}, replacing A with one of A, B, C, or D.

Your response should contain:
1. your reasoning
2. your final answer

Medical multiple-choice question:
{question}
\end{ICRPrompt}

\paragraph{ARC-C and GPQA-D.}
\begin{ICRPrompt}
You are an independent problem-solving agent.

Solve the multiple-choice question carefully and independently.

Reason from the evidence in the question.
Do not assume another agent will review your answer.

At the end, return exactly one final option in benchmark-compatible form: \boxed{A}, replacing A with one of A, B, C, or D.

Your response should contain:
1. your reasoning
2. your final answer

Multiple-choice question:
{question}
\end{ICRPrompt}

For GPQA-D, the following instruction is inserted immediately
after the sentence specifying the final option:

\begin{ICRPrompt}
Answer with a label from the final A-D list, not with the lowercase a)-d) items quoted inside those options.
\end{ICRPrompt}

\paragraph{GSM8K.}
\begin{ICRPrompt}
You are an independent problem-solving agent.

Solve the math word problem carefully and independently.

Show the reasoning needed to verify the calculation.
Do not assume another agent will review your answer.

At the end, return exactly one final numeric answer in the form \boxed{NUMBER}.

Math word problem:
{question}
\end{ICRPrompt}

\paragraph{HumanEval+.}
\begin{ICRPrompt}
You are an independent programming agent.

Solve the programming problem carefully and independently.
Check the function signature, edge cases, and examples in the problem.
Do not assume another agent will review your answer.

Return the complete implementation in exactly one markdown Python code block.

Programming problem:
{question}
\end{ICRPrompt}

For HumanEval+, the question field contains the following
loader-provided prefix followed by the benchmark prompt.
The displayed \texttt{\textbackslash n} sequences in this
prefix are literal characters.

\begin{ICRPrompt}
Please provide a self-contained Python script that solves the following problem in a markdown code block:\n```python\nYOUR_PYTHON_CODE\n```:
{benchmark_prompt}
\end{ICRPrompt}

\subsection{Receiver Revision Prompts}

For multiple-choice and numeric tasks, both receiver policies
use the following shared structure.
The receiver's prior fields remain fixed across communication
conditions.

\begin{ICRPrompt}
You previously solved this problem independently.

Original problem:
{question}

Your previous reasoning:
{receiver_prior_reasoning}

Your previous answer:
{receiver_prior_answer}

{external_block}

{policy_instructions}

{answer_format}
\end{ICRPrompt}

\paragraph{Critical Evaluation.}
The main audit uses the following policy instructions:

\begin{ICRPrompt}
Your task is to REVISE your belief, not to restart from scratch.

Evaluate your previous reasoning and any external message critically.

* Change your answer only if you find a concrete error in your previous reasoning, or evidence that is better supported than it.
* Do not change your answer merely because an external message is present.
* Resolve any disagreement using the evidence in the original problem.
\end{ICRPrompt}

\paragraph{Structured Verification.}
The receiver-policy comparison on MedQA and GPQA-D replaces
the preceding instructions with:

\begin{ICRPrompt}
Work through two steps, in this order.

Step 1 - Check the external message against the original problem.
Take each claim it makes and test it against the facts stated in the problem.
Do not compare it to your previous answer while doing this. Say which of its
claims hold and which do not. If no external message is available, say so and
go to Step 2.

Step 2 - Check your previous reasoning the same way, against the problem.
Say which of its claims hold and which do not.

Then give the answer supported by the claims that survived both steps. If the
surviving claims point to the answer you already gave, keep it. If they point
elsewhere, change it.
\end{ICRPrompt}

\paragraph{HumanEval+ revision.}
The supplementary code experiment uses a task-specific
Critical Evaluation template:

\begin{ICRPrompt}
You previously solved this problem independently.

Original problem:
{question}

Your previous reasoning and implementation:
{receiver_prior_reasoning}

{external_block}

Your task is to REVISE your implementation, not to restart from scratch and not to copy
an external message.

Evaluate your previous implementation and any external message critically.

* Change your implementation only if you find a concrete defect in it, or an approach that is better supported than it.
* Do not change your implementation merely because an external message is present.
* Check the required function signature, imports, examples, and edge cases against the original problem.

{answer_format}
\end{ICRPrompt}

\paragraph{External-information field.}
No Message supplies:

\begin{ICRPrompt}
No external message is available.
\end{ICRPrompt}

The other conditions use the following message block:

\begin{ICRPrompt}
An external message from another reasoning process is available.

External message:
{message_content}
\end{ICRPrompt}

For Answer Only, \texttt{message\_content} is the sender's
parsed answer converted to uppercase.
For Full Text, it is the sender's complete post-thinking response.
A missing parsed answer is represented as \texttt{UNPARSEABLE};
the same convention applies to the receiver's prior-answer field.

StateBridge and LatentMAS use the same visible message text
and the placeholder \texttt{[EMBEDDING\_CONTEXT\_HERE]}.
The placeholder is removed before tokenization.
StateBridge inserts its aligned embeddings at that position,
whereas LatentMAS supplies its payload as a KV prefix.
Appendix~B describes these delivery operations.

\paragraph{Task-specific output instructions.}
The following strings populate \texttt{answer\_format}
and are shared across receiver policies where applicable.

\noindent\textbf{MedQA}
\begin{ICRPrompt}
Return your concise reasoning, then exactly one final answer as \boxed{X}, where X is one of A, B, C, or D.
\end{ICRPrompt}

\noindent\textbf{ARC-C}
\begin{ICRPrompt}
Return your concise reasoning, then exactly one final answer as \boxed{X}, where X is one of the option labels shown above: a, b, c, or d.
\end{ICRPrompt}

\noindent\textbf{GPQA-D}
\begin{ICRPrompt}
Return your concise reasoning, then exactly one final answer as \boxed{X}, where X is one of A, B, C, or D from the final labeled list above. Do not answer with the lowercase a)-d) items quoted inside those options.
\end{ICRPrompt}

\noindent\textbf{GSM8K}
\begin{ICRPrompt}
Return your concise reasoning, then exactly one final answer as \boxed{N}, where N is a single number written in plain digits, with no thousands separators, no units, and no other symbols.
\end{ICRPrompt}

\noindent\textbf{HumanEval+}
\begin{ICRPrompt}
Return the complete final implementation in exactly one markdown Python code block. Do not put tests or explanatory prose inside that code block.
\end{ICRPrompt}

\subsection{Configuration and Prompt Verification}

Stored records include prompt hashes, configuration
fingerprints, per-record sampling seeds, generated token IDs,
and termination indicators.
Re-rendering the audited prompts from the recovered templates
reproduced their stored SHA-256 hashes.
Historical configuration fingerprints were also matched
to the corresponding generation limits.

These checks verify prompt reconstruction and configuration
mapping.
Exact source snapshots for every historical execution
were not archived, so a single commit does not fully
specify all evaluated runs.

\section{Communication Implementations}
\label{app:channels}

We implement each communication condition within the same
single-hop revision protocol, reusing the frozen initial
records across conditions.
Prompt templates are provided in Appendix~A.

\paragraph{Textual conditions.}
No Message supplies no sender information.
Answer Only supplies the sender's parsed final answer.
Full Text transmits the sender's complete post-thinking
response, including its expressed reasoning and final answer,
without additional generation or message shortening.
Textual messages are inserted into the external-information
slot of the receiver's revision prompt.

\paragraph{StateBridge.}
During independent generation, StateBridge records the output
of the last decoder block, before the final normalization,
at each decoding step.
The candidate sequence begins after the first generated
\texttt{</think>} token.
If this token is absent or no generated tokens follow it,
the complete generated sequence is used instead.
The last $K=\min(64,T)$ candidate states are retained
in their original order, where $T$ is the candidate length.
Short sequences are used without padding.

A whitened orthogonal Procrustes fit maps the selected states
to the input-embedding space using state--token-embedding
pairs from the same trajectory, with regularization $10^{-3}$.
The mapped states are rescaled to the mean embedding norm
and interpolated toward their nearest vocabulary embeddings
with weight $0.3$.
The resulting vectors are inserted at the external-message
slot in the receiver's input.

Each vector has dimension 2,560 and is stored in
\texttt{bfloat16}, giving a payload size of $5{,}120K$ bytes.
All retained StateBridge revisions on MedQA, ARC-C, and
GSM8K use $K=64$, corresponding to 327,680 bytes.
Two GPQA-D revisions and fourteen HumanEval+ revisions
use shorter payloads, with minimum lengths of 50 and
41 vectors, respectively.

\paragraph{LatentMAS.}
The LatentMAS adaptation reconstructs an all-layer KV cache
by teacher-forcing the sender's original prompt and complete
generated token sequence, including the thinking segment,
and then performs 10 latent steps.
The receiver accesses the resulting cache as a causal prefix
before its revision prompt.
StateBridge and LatentMAS use identical visible message-slot
text, while their payloads enter through different interfaces:
embedding insertion at the message slot versus KV-prefix
attachment before the prompt.

\paragraph{Comparison scope.}
These implementations differ in both the sender information
represented and its delivery.
Full Text transmits the post-thinking response;
StateBridge normally uses states from its final portion,
with the fallback described above;
LatentMAS provides a prefix derived from the full trajectory.
The audit therefore evaluates these message--interface
combinations under specified receiver policies, rather than
reproducing the methods' complete native agent workflows.
\section{Population, Scoring, and Statistical Scope}
\label{app:scoring}

\paragraph{Dataset sources.}
MedQA uses all entries in a fixed local file containing
300 questions; the original subset-selection procedure
and source-question identifiers are unavailable.
ARC-C uses the \texttt{ARC-Challenge} test split of
\texttt{allenai/ai2\_arc}.
GSM8K uses the \texttt{main} test split of
\texttt{openai/gsm8k}.
GPQA-D uses a local conversion of all 198 Diamond questions,
with options reordered and labeled A--D.
The original option-permutation procedure is unavailable.
HumanEval+ uses the 164-item test split of
\texttt{evalplus/humanevalplus}.

Questions retain their file or dataset order.
For the three datasets loaded from Hugging Face,
reconstructed loader outputs match the dataset hashes
stored in the experiment configurations.

\input{tables/patterns}

\paragraph{Selection.}
ARC-C excludes seven items whose option count is not four
(IDs 121, 385, 400, 836, 868, 1037, and 1042).
For each task, the retained population consists of questions
with at least one initially incorrect agent.
Selection and stratification use initial correctness only;
reference answers and correctness labels are not provided
to message construction or receiver revision.

For a triple with $k$ correct answers, complete directed pairing gives $k(3-k)$ CR directions, the same number of PR directions, $(3-k)(2-k)$ SR directions, and $k(k-1)$ SCR directions. These counts reproduce the completed-task strata. The four main-text datasets contain 430 retained questions
and 2,580 directions per condition.
HumanEval+ adds 28 retained questions and 168 directions
per condition.

\paragraph{MedQA evaluation scope.}
The retained MedQA population contains 720 directions:
116 each for CR and PR, 436 for SR, and 52 for SCR.
Retained-subset accuracy is computed exclusively on these
directions.
Outcomes from all-three-correct questions are handled
separately in the full-set estimates described in Appendix~\ref{app:extrapolation}.

\paragraph{Parsing and scoring.}
For multiple-choice tasks, the parser uses the last
\verb|\boxed{...}| expression, removes LaTeX formatting,
and extracts a valid option label.
Labels are compared case-insensitively.
Responses without a valid boxed label are unparsed;
earlier boxed expressions are not used as fallbacks.

For GSM8K, the parser extracts the first number from the
last boxed expression.
If that expression contains no number, its cleaned text
is retained.
When no boxed expression is present, the parser uses the
last number in the response.
Thousands separators are removed, and numeric answers
are compared as exact decimal values without a tolerance.
If either value is nonnumeric, comparison uses string equality.

For HumanEval+, the parser extracts the last fenced code
block tagged as Python.
The implementation is executed with the dataset's
\texttt{test} field in a fresh subprocess.
A response passes if the complete program exits successfully
within a 10-second wall-clock limit covering all tests
for that question.
The official EvalPlus evaluation harness is not used.

Unparsed outputs are scored incorrect.
Non-terminating outputs remain in the evaluation and are
scored using the same task-specific rules.
Termination status and parsing failure are therefore
reported separately.
\input{tables/termination}
\input{tables/truncation_sensitivity}

\input{tables/raw_integrity}

The all-correct observations cover only the A-to-B and
B-to-A directions on MedQA, ARC-C, GSM8K, and HumanEval+.
On GPQA-D, they comprise partial directional coverage
from an interrupted collection run.
Their use in full-set estimation is described in Appendix~G.

\paragraph{Sensitivity to initial truncation.}
On GPQA-D, excluding the nine questions with at least one
truncated initial trajectory reduces the CR and PR
denominators from 114 to 100 each.
SI decreases for all four conditions.
The ordering among communication conditions remains unchanged,
while LatentMAS falls below No Message.
This sensitivity analysis uses a filtered population;
the main results retain all questions.

\paragraph{Evaluation scope.}
No separate question-level development split is documented
in the experiment records.
Generation budgets were adjusted during data collection
as described in Appendix~A.
The results characterize the evaluated configurations
on the reported question populations.

\subsection{Uncertainty Estimates}
\label{app:uncertainty}

Marginal intervals for retained accuracy, CR, and PR use
2,000 question-cluster bootstrap resamples with analysis
seed 20260921 and percentile 2.5/97.5 bounds.
All directions associated with a resampled question
are retained together.

\input{tables/intervals}

The message-content and receiver-policy difference analyses
use 10,000 paired question-cluster bootstrap resamples.
Compared conditions are resampled jointly by question,
preserving event correspondence and within-question
dependence.
Difference intervals are computed directly from paired
outcomes, rather than reconstructed from marginal bounds.

These intervals are conditional on the evaluated runs
and question population and do not capture variation
across unobserved generation seeds.
We do not infer significance from overlap between marginal
intervals or treat the six directions as independent
observations.
Intervals containing zero do not establish equivalence.

\section{Paired Outcomes and Answer Transitions}
\label{app:paired}

\input{tables/paired}

Each communication condition is paired with No Message
by question and ordered sender--receiver pair, with identical
receiver priors.
A gain occurs when revision is correct with communication
but incorrect without it; a loss is the reverse.
The net change in correct outcomes is the number of gains
minus losses.
Dividing this quantity by the number of paired events gives
the retained-subset accuracy difference.

These comparisons include all four initial-correctness
strata.
They therefore capture changes in both mixed-correctness
revision and the recovery or preservation of answers when
both agents are initially wrong or correct.
Gains and losses are defined by correctness, independently
of answer identity.

\subsection{Answer Identity}
\label{app:transitions}

\input{tables/transitions}

\paragraph{Normalization.}
For choice and numeric tasks, identity comparisons use
parsed answer strings after trimming leading and trailing
whitespace and converting to lowercase.
For GSM8K, this is string identity rather than numerical
equality: \texttt{18} and \texttt{18.0} would count as
different answers, although correctness scoring treats them
as equal.
No such discrepancy occurs in the reported GSM8K records.
For HumanEval+, the reported counts are reproduced by
comparing extracted program strings after trimming leading
and trailing whitespace.
Program-string identity does not imply functional equivalence.

\paragraph{Category precedence.}
The four categories are mutually exclusive and exhaustive,
with the following precedence:
\emph{Invalid} if the revised answer cannot be parsed;
otherwise \emph{Prior} if it matches the receiver's initial
answer;
otherwise \emph{Sender} if it matches the sender's initial
answer;
otherwise \emph{Third}.
When a revised answer matches both initial answers,
it is therefore counted as Prior.
The Sender category counts only revisions that differ
from the receiver's initial answer.

Records with unparsed initial answers are retained.
A valid revised answer cannot match an unparsed receiver
prior.
If the sender's initial answer is unparsed, a valid revision
that differs from the receiver's prior is included in Third.
Thus, Third also includes changes for which no parsed
sender answer is available.

\paragraph{Interpretation.}
Sender matching describes an observed answer relationship,
not causal adoption of a message.
No Message can also produce a sender-matching answer.
Likewise, an answer assigned to Third may be correct or
incorrect.
Answer identity and correctness are therefore analyzed
separately.

\paragraph{Valid-output change rate.}
A revised output is valid if its answer can be parsed.
The valid-output change rate is the fraction of valid
revised answers that differ from the receiver's initial
answer.
Invalid revised outputs are excluded from both the numerator
and denominator.
Records with unparsed initial answers remain included;
a valid revision from an unparsed receiver prior counts
as a change.
When combining CR and PR strata, we sum the corresponding
numerators and denominators before computing the rate.

\paragraph{Answer Only and Full Text.}
Table~\ref{tab:ao-ft-transitions} reports sender-matching
counts and valid-output change rates on the shared
mixed-correctness directions.
Full Text produces more sender-matching revisions in both
strata and a higher pooled valid-output change rate on all
four datasets.
These results characterize more frequent answer revision;
the correctness-conditioned metrics determine whether
those revisions correct or corrupt the receiver's answer.

\begin{table}[t]
\centering
\caption{
Answer Only and Full Text on shared mixed-correctness
directions under Critical Evaluation.
Sender matches report counts over all events in each stratum,
using Prior-before-Sender precedence.
Valid change rates pool CR and PR and exclude unparsed
revised outputs.
Each GPQA-D condition has one unparsed revised output;
the other conditions shown have none.
}
\small
\setlength{\tabcolsep}{4pt}
\renewcommand{\arraystretch}{1.08}
\begin{tabular}{@{}llrrl@{}}
\toprule
& & \multicolumn{2}{c}{Sender matches}
& \\
\cmidrule(lr){3-4}
Dataset & Condition & CR & PR & Valid change rate \\
\midrule
MedQA
& Answer Only & 45/116 & 23/116 & 68/232 (29.31\%) \\
& Full Text   & 93/116 & 67/116 & 160/232 (68.97\%) \\
\midrule
ARC-C
& Answer Only & 38/98 & 36/98 & 74/196 (37.76\%) \\
& Full Text   & 79/98 & 64/98 & 143/196 (72.96\%) \\
\midrule
GSM8K
& Answer Only & 36/92 & 23/92 & 64/184 (34.78\%) \\
& Full Text   & 47/92 & 43/92 & 93/184 (50.54\%) \\
\midrule
GPQA-D
& Answer Only & 64/114 & 38/114 & 107/227 (47.14\%) \\
& Full Text   & 85/114 & 58/114 & 144/227 (63.44\%) \\
\bottomrule
\end{tabular}

\label{tab:ao-ft-transitions}
\end{table}

\section{Receiver-Side Cost Observations}
\label{app:cost}
\input{tables/cost}

The current export reports zero additional construction generation calls for all four baselines, but that field does not measure forward-pass computation. LatentMAS reconstructs the sender cache and runs latent steps; StateBridge performs alignment whose timing is stored with the initial belief. Message-construction time is exported only for LatentMAS (mean 0.483, 0.393, 0.547, and 0.721 seconds on MedQA, ARC-C, GSM8K, and HumanEval+, respectively). These fields do not establish a comparable end-to-end cost across channels.

The LatentMAS payload column is zero throughout the exported cost CSV despite the documented KV cache. We treat its byte size as unavailable in the paper rather than report zero communication. Logged receiver-generation times also reflect different worker concurrency and contention. They are descriptive measurements of these runs, not a controlled speedup or throughput comparison.

\section{HumanEval+ Supplement}
\label{app:humaneval}
\input{tables/humaneval}

HumanEval+ has 28 CR and 28 PR directions, so one outcome changes either rate by 3.57 percentage points. Nine directions in each stratum involve a non-terminating initial trajectory. The audit identifies 12 such initial records across eight questions, not 12 excluded questions. It also records 49 non-terminating revision records across ten questions in the full raw revision scope. No HumanEval+ item exclusion was applied, and all conditions share the same 168 retained directions.

No Message already corrects 16/28 eligible receiver answers. Full Text increases this to 21/28 without changing PR, while StateBridge obtains 19/28 CR and 27/28 PR. Thus, the main-text statement that all communication channels reduce PR relative to No Message does not extend to this supplementary dataset. LatentMAS has the highest CR (24/28) but a lower PR (17/28), and its SI is below No Message. These observations are limited by the small sample and generation failures. 

\paragraph{Execution-based scoring.}
Programs are evaluated against the dataset-provided test
code with a 10-second wall-clock limit per question,
without the official EvalPlus evaluation harness.
A post-hoc serial re-execution produced different correctness
outcomes for one question.
The reported results retain the original execution scores
and the initial-correctness strata defined from them.

\section{Estimated Full-Set Accuracy}
\label{app:extrapolation}

\input{tables/extrapolation}

Full-set accuracy combines observed revision outcomes
with assumptions about unobserved directions.
The observation scope differs between Qwen3-4B and Qwen3-8B,
as specified below.
For both models, $N$ denotes the number of evaluated
questions before correctness-based selection, giving
$6N$ ordered sender--receiver events.

\paragraph{Qwen3-4B estimates.}
For condition $c$, let $C_{\rm ret}^{(c)}$ denote correct
retained outcomes, $C_{\rm samp}^{(c)}$ correct observed
outcomes from all-three-correct questions, and
$n_{\rm miss}^{(c)}$ the number of unobserved directions.
The estimate is
\begin{equation}
\widehat{\mathrm{Acc}}_{\rm full}^{(c)}
=
\frac{
C_{\rm ret}^{(c)}
+
C_{\rm samp}^{(c)}
+
n_{\rm miss}^{(c)}\widehat p_{\rm skip}^{(c)}
}{6N}.
\end{equation}
The retained, observed all-three-correct, and unobserved
directions form disjoint parts of the evaluation population.
Where all-three-correct observations are available,
$\widehat p_{\rm skip}^{(c)}$ is their empirical accuracy.

\paragraph{Qwen3-8B estimates.}
All five conditions are evaluated only on mixed-correctness
directions.
For every unobserved direction, we assume that revision
preserves the receiver's initial correctness:
initially correct receivers remain correct, and initially
incorrect receivers remain incorrect.
Let $\mathcal{M}$ denote the observed mixed-correctness
directions and $\mathcal{U}$ the unobserved directions.
The estimate is
\begin{equation}
\widehat{\mathrm{Acc}}_{\rm full}^{(c)}
=
\frac{
C_{\rm mixed}^{(c)}
+
\sum_{e\in\mathcal{U}} b_R(e)
}{6N},
\end{equation}
where $C_{\rm mixed}^{(c)}$ counts correct revised answers
on $\mathcal{M}$, and $b_R(e)$ indicates whether the
receiver's initial answer is correct.

MedQA has 184 observed and 1,616 unobserved directions,
of which 1,328 unobserved directions have an initially
correct receiver.
GPQA-D has 216 observed and 972 unobserved directions,
of which 580 unobserved directions have an initially
correct receiver.
The corresponding estimates are therefore
$(C_{\rm mixed}^{(c)}+1{,}328)/1{,}800$ for MedQA and
$(C_{\rm mixed}^{(c)}+580)/1{,}188$ for GPQA-D.

\paragraph{Observed coverage.}
On MedQA, ARC-C, GSM8K, and HumanEval+, the available
all-three-correct observations cover the A-to-B and B-to-A
directions collected during an earlier two-agent stage.
Only questions for which all three cached initial answers
are correct contribute to this population.
GPQA-D observations provide partial directional coverage
from an interrupted collection run.
Observed sets are not fully matched across conditions:
GSM8K has 235 observations for No Message, Full Text,
and StateBridge, but 233 for LatentMAS.
Coverage denotes the fraction of all directions with
observed revision outcomes, including both retained
and all-three-correct observations.

\paragraph{Interpretation.}
The Qwen3-4B estimates generally extrapolate observed
all-three-correct accuracy to unobserved directions;
the Qwen3-8B estimates instead assume unchanged correctness
on all directions outside the mixed-correctness population.
These assumptions are distinct, despite the shared
estimated-accuracy notation in the main table.
The available all-three-correct observations do not
constitute a documented probability sample of omitted
directions.
We report point estimates without confidence intervals
and distinguish them from directly measured outcomes.

\section{Six-Direction Detail}
\label{app:directions}
Table~\ref{tab:directions} preserves the six ordered directions for all completed datasets. These are correlated views of the same questions and should not be treated as six independent replications.
\input{tables/directions}

\section{Supplementary Controlled Comparisons}
\label{app:controlled_comparisons}

\subsection{Random-Delivery Reference}
\label{app:random_delivery}

We construct a content-independent delivery reference using
the existing Full Text and No Message outcomes on the same
mixed-correctness directions.
Within each dataset, the reference supplies Full Text with
probability $\alpha$ and No Message otherwise, independently
of the question and initial correctness.
Its expected correction and preservation rates are
\begin{align}
\mathrm{CR}_{\alpha}
&=
\alpha\,\mathrm{CR}_{\mathrm{Full}}
+(1-\alpha)\,\mathrm{CR}_{\mathrm{None}},
\label{eq:random_delivery_cr}\\
\mathrm{PR}_{\alpha}
&=
\alpha\,\mathrm{PR}_{\mathrm{Full}}
+(1-\alpha)\,\mathrm{PR}_{\mathrm{None}}.
\label{eq:random_delivery_pr}
\end{align}
Here, Full and None denote Full Text and No Message.
These expected rates are computed from existing outcomes;
the reference does not require additional model generation.

For each dataset, we choose
\begin{equation}
\alpha
=
\frac{\mathrm{PR}_{\mathrm{None}}
      -\mathrm{PR}_{\mathrm{AnswerOnly}}}
     {\mathrm{PR}_{\mathrm{None}}
      -\mathrm{PR}_{\mathrm{Full}}},
\label{eq:random_delivery_alpha}
\end{equation}
so that $\mathrm{PR}_{\alpha}$ matches Answer Only's PR.
All four fitted probabilities lie in $[0,1]$.
We then report
$\Delta\mathrm{CR}
=\mathrm{CR}_{\mathrm{AnswerOnly}}-\mathrm{CR}_{\alpha}$:
positive values indicate more correction by Answer Only
at the matched preservation rate.

\begin{table}[t]
\centering
\caption{
Answer Only minus the random-delivery reference at matched PR.
Reference CR is a percentage; differences and intervals
are in percentage points.
95\% CIs use 10,000 paired question-cluster bootstrap
resamples.
}
\small
\setlength{\tabcolsep}{6pt}
\renewcommand{\arraystretch}{1.08}
\begin{tabular}{@{}lrrrr@{}}
\toprule
Dataset
& $\alpha$
& Ref.\ CR (\%)
& $\Delta$CR
& 95\% CI \\
\midrule
MedQA
& $0.3231$ & $30.57$ & $+8.22$
& $[-2.99,\;19.86]$ \\
ARC-C
& $0.5088$ & $45.02$ & $-6.25$
& $[-19.78,\;7.86]$ \\
GSM8K
& $0.5405$ & $33.61$ & $+5.52$
& $[-9.85,\;19.01]$ \\
GPQA-D
& $0.6786$ & $55.11$ & $+1.03$
& $[-14.52,\;15.57]$ \\
\bottomrule
\end{tabular}

\label{tab:random_delivery_reference}
\end{table}

Table~\ref{tab:random_delivery_reference} provides the
numerical results underlying the matched-PR comparison.
Point estimates favor Answer Only on three datasets and
the reference on ARC-C.
All four intervals include zero; these comparisons do not
establish equivalence or identify reduced message exposure
as the mechanism underlying Answer Only's behavior.

\subsection{Channel Comparisons Across Receiver Policies}
\label{app:channel_policy_comparison}

We examine how the SI contrast between StateBridge and
Full Text changes with the receiver policy.
Within each channel, the comparison reuses the initial
trajectories, sender messages, and delivery interface.
MedQA and GPQA-D contain 116 and 114 events per
mixed-correctness stratum, respectively.

Let $p_{\mathrm{CE}}$ denote Critical Evaluation and
$p_{\mathrm{SV}}$ denote Structured Verification.
For each policy, define the channel contrast
\begin{equation}
D(p)
=
\mathrm{SI}(\mathrm{StateBridge},p)
-
\mathrm{SI}(\mathrm{FullText},p).
\end{equation}
The channel-by-policy interaction is
\begin{equation}
I=D(p_{\mathrm{SV}})-D(p_{\mathrm{CE}}).
\end{equation}
A positive $I$ means that moving to Structured Verification
shifts the SI contrast toward StateBridge; a negative value
means that it shifts toward Full Text.
This interaction concerns relative channel performance,
rather than either channel's policy effect in isolation.

\begin{table}[t]
\centering
\caption{
StateBridge minus Full Text in SI under each receiver
policy, and the change in this contrast.
All values are in percentage points.
Contrasts and interactions are computed before rounding.
95\% CIs use 10,000 paired question-cluster bootstrap
resamples.
}
\small
\setlength{\tabcolsep}{6pt}
\renewcommand{\arraystretch}{1.08}
\begin{tabular}{@{}lrrrr@{}}
\toprule
Dataset
& $D(p_{\mathrm{CE}})$
& $D(p_{\mathrm{SV}})$
& $I$
& 95\% CI for $I$ \\
\midrule
MedQA
& $+1.72$ & $-5.60$ & $-7.33$
& $[-15.09,\;0.00]$ \\
GPQA-D
& $-3.07$ & $+2.19$ & $+5.26$
& $[-4.39,\;14.91]$ \\
\bottomrule
\end{tabular}

\label{tab:channel_policy_interaction}
\end{table}

Bootstrap resampling operates on questions, retaining
their associated directions and compared conditions
together.
The point-estimate channel ordering changes in opposite
directions across the two datasets.
Both interaction intervals include zero, so these
descriptive ordering changes do not establish a
channel-by-policy interaction on either dataset.

\subsection{Correction and Preservation Differences}
\label{app:message_metric_differences}

We compare Full Text with Answer Only on their shared
mixed-correctness directions.
Table~\ref{tab:ft-minus-ao} reports paired differences
in CR and PR.
Full Text yields higher CR and lower PR on all four datasets.
The CR difference intervals exclude zero on MedQA, ARC-C,
and GPQA-D, while the GSM8K interval includes zero.
All four PR difference intervals lie below zero.

\begin{table}[t]
\centering
\small
\caption{
Full Text minus Answer Only on shared mixed-correctness
directions. Differences and intervals are in percentage
points. The 95\% CIs use 10,000 paired question-cluster
bootstrap resamples, retaining associated directions
and compared conditions together.
}
\setlength{\tabcolsep}{4pt}
\renewcommand{\arraystretch}{1.08}
\begin{tabular}{@{}lrrrrr@{}}
\toprule
Dataset
& $n_{\mathrm{CR}}/n_{\mathrm{PR}}$
& $\Delta$CR
& 95\% CI
& $\Delta$PR
& 95\% CI \\
\midrule
MedQA
& 116/116
& $+41.38$ & $[30.17,52.59]$
& $-37.93$ & $[-49.14,-26.72]$ \\
ARC-C
& 98/98
& $+41.84$ & $[28.57,54.08]$
& $-28.57$ & $[-40.82,-16.33]$ \\
GSM8K
& 92/92
& $+11.96$ & $[0.00,23.91]$
& $-18.48$ & $[-31.52,-5.43]$ \\
GPQA-D
& 114/114
& $+18.42$ & $[8.77,28.07]$
& $-15.79$ & $[-26.32,-6.14]$ \\
\bottomrule
\end{tabular}

\label{tab:ft-minus-ao}
\end{table}

\subsection{Policy-Specific Communication Increments}
\label{app:policy_communication_increments}

We define the communication increment as
$G(c,p)=\mathrm{SI}(c,p)-\mathrm{SI}(\text{No Message},p)$.
Its change across receiver policies is
$\Delta G(c)=G(c,\mathrm{SV})-G(c,\mathrm{CE})$,
where CE and SV denote Critical Evaluation and
Structured Verification, respectively.
This contrast measures how the channel's SI advantage
over no-message revision changes with receiver policy.

Table~\ref{tab:receiver-policy-gain} reports all three
communication conditions on both datasets.
Bootstrap resampling jointly retains the channel and
No Message outcomes under both policies, together with
all associated directions for each question.
Contrasts are computed before rounding.

\begin{table}[t]
\centering
\small
\caption{
Communication increments relative to the policy-specific
No Message control and their changes from CE to SV.
All values are in percentage points.
The 95\% CIs use 10,000 paired question-cluster bootstrap
resamples, jointly resampling all four condition--policy cells.
}
\setlength{\tabcolsep}{5pt}
\renewcommand{\arraystretch}{1.08}
\begin{tabular}{@{}llrrrr@{}}
\toprule
Dataset & Channel
& $G(c,\mathrm{CE})$
& $G(c,\mathrm{SV})$
& $\Delta G$
& 95\% CI for $\Delta G$ \\
\midrule
MedQA
& Full Text
& $+8.62$ & $+10.78$ & $+2.16$
& $[-3.88,8.19]$ \\
& StateBridge
& $+10.34$ & $+5.17$ & $-5.17$
& $[-11.21,0.86]$ \\
& LatentMAS
& $-1.29$ & $+9.91$ & $+11.21$
& $[2.16,20.26]$ \\
\midrule
GPQA-D
& Full Text
& $+5.70$ & $+6.14$ & $+0.44$
& $[-7.89,8.77]$ \\
& StateBridge
& $+2.63$ & $+8.33$ & $+5.70$
& $[-2.19,13.60]$ \\
& LatentMAS
& $+0.88$ & $+7.02$ & $+6.14$
& $[-2.63,15.35]$ \\
\bottomrule
\end{tabular}

\label{tab:receiver-policy-gain}
\end{table}

On MedQA, the LatentMAS increase in communication increment
has a strictly positive interval.
The remaining $\Delta G$ intervals include zero.

\subsection{Sensitivity to Revision Truncation}
\label{app:policy_truncation}

We assess whether the GPQA-D receiver-policy comparisons
are sensitive to truncated revision outputs.
The analysis covers No Message, Full Text, and StateBridge
under Critical Evaluation and Structured Verification.
We apply a common exclusion mask across all six
condition--policy combinations: a directed event is removed
if its revision output is truncated in any combination.
This removes 6 of the 228 shared events, leaving 111 events
in each of the CR and PR strata.

Table~\ref{tab:policy_truncation} reports policy differences
on this common subset.
All contrasts are computed as Structured Verification
minus Critical Evaluation.
Communication increments use the corresponding
No Message outcomes on the same subset.
The interaction is the change in the StateBridge-minus-Full
Text SI contrast across policies.

\begin{table}[t]
\centering
\small
\caption{
GPQA-D receiver-policy contrasts after jointly excluding
events with a truncated revision in any of the six
evaluated condition--policy combinations.
The common subset contains 111 CR and 111 PR events.
All values are in percentage points.
Intervals use paired question-cluster bootstrap resampling,
retaining associated directions and compared conditions
together.
}
\setlength{\tabcolsep}{6pt}
\renewcommand{\arraystretch}{1.08}
\begin{tabular}{@{}llrr@{}}
\toprule
Contrast & Condition & Estimate & 95\% CI \\
\midrule
$\Delta\mathrm{SI}$
& No Message
& $-1.35$ & $[-4.91,\;2.23]$ \\
& Full Text
& $-3.15$ & $[-9.95,\;3.60]$ \\
& StateBridge
& $+2.25$ & $[-4.68,\;9.09]$ \\
\midrule
$\Delta G$
& Full Text
& $-1.80$ & $[-9.91,\;6.25]$ \\
& StateBridge
& $+3.60$ & $[-3.90,\;11.24]$ \\
\midrule
Interaction $I$
& StateBridge versus Full Text
& $+5.41$ & $[-4.51,\;15.39]$ \\
\bottomrule
\end{tabular}

\label{tab:policy_truncation}
\end{table}

On the common subset, the SI point estimates decrease
for No Message and Full Text and increase for StateBridge.
All reported contrast intervals include zero.
Because exclusion depends on post-revision termination,
this analysis describes a selected subset.
The main analysis retains all events and applies the
task-specific scoring rules irrespective of termination status.

\section{Additional Results with Qwen3-8B}
\label{app:qwen3_8b}

\paragraph{Evaluation setup.}
We evaluate \texttt{Qwen/Qwen3-8B} on MedQA and GPQA-D
using Critical Evaluation and a generation limit of
16,384 new tokens.
Three agents independently solve each question.
Their initial trajectories are frozen and reused across
No Message, Answer Only, Full Text, StateBridge, and LatentMAS.
Revision is evaluated only on mixed-correctness directions.
The initial trajectories and resulting audit populations
are specific to each model size.

\paragraph{Initial correctness and coverage.}
Table~\ref{tab:8b_population} summarizes the initial
three-agent correctness patterns.
MedQA contains 46 mixed-correctness questions, yielding
92 CR and 92 PR events per condition.
GPQA-D contains 54 such questions, yielding 108 events
per stratum.
All five conditions have complete coverage of these events,
with no duplicate directional keys and matched receiver
priors across conditions.
SR and SCR are not measured for Qwen3-8B.
Full-set accuracy estimates use the assumptions described
in Appendix~\ref{app:extrapolation}.

\begin{table}[t]
\centering
\caption{
Qwen3-8B initial correctness.
$k$ is the number of initially correct agents.
Initial accuracy is measured over all three agents;
questions with $k=1$ or $k=2$ contribute mixed-correctness
directions.
}
\small
\setlength{\tabcolsep}{5pt}
\renewcommand{\arraystretch}{1.08}
\begin{tabular}{@{}lrrrrrr@{}}
\toprule
Dataset & Questions
& $k=0$ & $k=1$ & $k=2$ & $k=3$
& Initial acc.\ (\%) \\
\midrule
MedQA  & 300 & 40 & 24 & 22 & 214 & 78.89 \\
GPQA-D & 198 & 57 & 25 & 29 & 87  & 57.91 \\
\bottomrule
\end{tabular}

\label{tab:8b_population}
\end{table}

\paragraph{Correction and preservation.}
Table~\ref{tab:8b_audit} reports the conditional rates
and their marginal intervals.
On both tasks, Full Text has higher CR and lower PR
than Answer Only.
StateBridge has the highest observed SI on MedQA,
whereas Answer Only has the highest SI point estimate
on GPQA-D.
The relative profiles therefore depend on the task.

\begin{table}[t]
\centering
\caption{
Qwen3-8B audit results under Critical Evaluation (\%).
CR and PR denominators are 92 each on MedQA and
108 each on GPQA-D.
Marginal 95\% intervals use 10,000 question-cluster
bootstrap resamples.
Bold marks the highest SI point estimate per dataset.
}
\small
\setlength{\tabcolsep}{4pt}
\renewcommand{\arraystretch}{1.08}
\begin{tabular}{@{}llrrr@{}}
\toprule
Dataset & Condition
& CR [95\% CI]
& PR [95\% CI]
& SI \\
\midrule
MedQA
& No Message
& 7.61 [1.35, 15.12]
& 93.48 [86.90, 98.72]
& 50.54 \\
& Answer Only
& 31.52 [20.54, 42.86]
& 85.87 [76.32, 94.23]
& 58.70 \\
& Full Text
& 77.17 [66.04, 86.96]
& 46.74 [33.70, 60.47]
& 61.96 \\
& StateBridge
& 60.87 [48.89, 72.73]
& 76.09 [66.67, 84.91]
& \textbf{68.48} \\
& LatentMAS
& 73.91 [63.54, 83.72]
& 41.30 [30.30, 52.56]
& 57.61 \\
\midrule
GPQA-D
& No Message
& 14.81 [7.55, 22.92]
& 92.59 [86.76, 97.46]
& 53.70 \\
& Answer Only
& 50.93 [40.00, 61.76]
& 65.74 [55.00, 76.09]
& \textbf{58.33} \\
& Full Text
& 70.37 [60.20, 80.16]
& 44.44 [33.00, 56.16]
& 57.41 \\
& StateBridge
& 65.74 [54.65, 76.25]
& 46.30 [34.91, 57.76]
& 56.02 \\
& LatentMAS
& 44.44 [32.98, 55.68]
& 67.59 [57.46, 77.27]
& 56.02 \\
\bottomrule
\end{tabular}

\label{tab:8b_audit}
\end{table}

\paragraph{Paired communication increments.}
Table~\ref{tab:8b_gain} compares each communication condition
with No Message on the same directed events.
On MedQA, the SI increments for Answer Only, Full Text,
and StateBridge have intervals above zero.
The LatentMAS interval includes zero.
On GPQA-D, all four increment intervals include zero.
These contrasts are assessed directly from paired outcomes,
rather than from overlap between marginal intervals.

\begin{table}[t]
\centering
\caption{
Qwen3-8B SI differences relative to No Message,
in percentage points.
Intervals use 10,000 paired question-cluster bootstrap
resamples, preserving associated directions and
compared conditions together.
Differences are computed before rounding.
}
\small
\setlength{\tabcolsep}{6pt}
\renewcommand{\arraystretch}{1.08}
\begin{tabular}{@{}llrr@{}}
\toprule
Dataset & Condition & $\Delta$SI & 95\% CI \\
\midrule
MedQA
& Answer Only & $+8.15$  & $[2.17,14.39]$ \\
& Full Text   & $+11.41$ & $[1.97,20.75]$ \\
& StateBridge & $+17.93$ & $[9.90,25.71]$ \\
& LatentMAS   & $+7.07$  & $[-1.92,16.07]$ \\
\midrule
GPQA-D
& Answer Only & $+4.63$ & $[-2.31,11.57]$ \\
& Full Text   & $+3.70$ & $[-4.41,12.07]$ \\
& StateBridge & $+2.31$ & $[-5.70,10.17]$ \\
& LatentMAS   & $+2.31$ & $[-6.07,10.50]$ \\
\bottomrule
\end{tabular}

\label{tab:8b_gain}
\end{table}

StateBridge also exceeds Answer Only in SI on MedQA
by 9.78 percentage points
(95\% CI: $[0.66,18.62]$).
The corresponding GPQA-D difference is
$-2.31$ points ($[-9.84,5.24]$).
These comparisons concern within-model communication
conditions; the model-specific audit populations are
not treated as paired samples across model sizes.

\paragraph{Termination.}
Independent generation contains one truncated response
on MedQA and seven on GPQA-D.
On MedQA, revision truncation occurs in two No Message,
two Answer Only, one Full Text, and one LatentMAS output,
all associated with the same question.
On GPQA-D, one Answer Only and one StateBridge output
are truncated, also on the same question.
All events remain in the reported analysis and are scored
using the task-specific parsing and correctness rules.

%% file: tables/patterns.tex
\begin{table}[t]
\centering
\caption{Initial three-agent correctness patterns and
initial-answer accuracy (\%).
$k$ denotes the number of initially correct agents.
Initial accuracy is pooled over the three agents.
ARC-C counts reflect the seven structural exclusions.
GPQA-D phase-1 statistics are complete.}
\small
\setlength{\tabcolsep}{4pt}
\begin{tabular}{lrrrrrr}
\toprule
Dataset & Items & $k=3$ & $k=2$ & $k=1$ & $k=0$ & Initial acc. \\
\midrule
MedQA & 300 & 180 & 26 & 32 & 62 & 69.33 \\
ARC-C & 1165 & 1067 & 32 & 17 & 49 & 93.91 \\
GSM8K & 1319 & 1225 & 23 & 23 & 48 & 94.62 \\
GPQA-D & 198 & 80 & 24 & 33 & 61 & 54.04 \\
HumanEval+ & 164 & 136 & 10 & 4 & 14 & 87.80 \\
\bottomrule
\end{tabular}

\label{tab:patterns}
\end{table}

%% file: tables/termination.tex
\begin{table}[t]
\centering
\small
\caption{Directions involving at least one initial trajectory flagged as non-terminating. This flag is not identical to an unparsed or semantically incorrect answer; GSM8K contains non-EOS trajectories with parsed answers.}
\setlength{\tabcolsep}{4pt}
\begin{tabular}{lrrrr}
\toprule
Dataset & CR & PR & SR & SCR \\
\midrule
MedQA & 0/116 (0.00\%) & 0/116 (0.00\%) & 4/436 (0.92\%) & 0/52 (0.00\%) \\
ARC-C & 2/98 (2.04\%) & 2/98 (2.04\%) & 4/328 (1.22\%) & 0/64 (0.00\%) \\
GSM8K & 4/92 (4.35\%) & 4/92 (4.35\%) & 0/334 (0.00\%) & 0/46 (0.00\%) \\
GPQA-D
& 12/114 (10.53\%)
& 12/114 (10.53\%)
& 18/432 (4.17\%)
& 0/48 (0.00\%) \\
HumanEval+ & 9/28 (32.14\%) & 9/28 (32.14\%) & 20/92 (21.74\%) & 0/20 (0.00\%) \\
\bottomrule
\end{tabular}

\label{tab:termination}
\end{table}

%% file: tables/truncation_sensitivity.tex
\begin{table}[t]
\centering
\small
\caption{GPQA-D sensitivity to excluding the nine questions
with at least one truncated initial trajectory.
All and Excl.\ denote the original and filtered populations.
CR and PR denominators decrease from 114 to 100 each.
Metrics are percentages; $\Delta$SI is in percentage points.
The main results retain all questions.}
\setlength{\tabcolsep}{4pt}
\renewcommand{\arraystretch}{1.08}
\begin{tabular}{@{}lrrrrrrr@{}}
\toprule
& \multicolumn{2}{c}{CR}
& \multicolumn{2}{c}{PR}
& \multicolumn{2}{c}{SI}
& \\
\cmidrule(lr){2-3}
\cmidrule(lr){4-5}
\cmidrule(lr){6-7}
Method
& All & Excl.
& All & Excl.
& All & Excl.
& $\Delta$SI \\
\midrule
No Message
& 14.04 & 9.00
& 96.49 & 96.00
& 55.26 & 52.50
& -2.76 \\
Full Text
& 74.56 & 72.00
& 47.37 & 41.00
& 60.96 & 56.50
& -4.46 \\
StateBridge
& 64.04 & 61.00
& 51.75 & 47.00
& 57.89 & 54.00
& -3.89 \\
LatentMAS
& 61.40 & 56.00
& 50.88 & 48.00
& 56.14 & 52.00
& -4.14 \\
\bottomrule
\end{tabular}

\label{tab:truncation_sensitivity}
\end{table}

%% file: tables/raw_integrity.tex
\begin{table}[t]
\centering
\small
\caption{
Record counts, non-termination, and parsing failures.
Revision counts pool No Message, Full Text, StateBridge,
and LatentMAS under Critical Evaluation.
Retained and observed all-three-correct questions form
disjoint populations; imputed outcomes are excluded.
No all-correct observations are available for the reported
MedQA StateBridge condition.
No EOS and Unparsed are overlapping indicators,
not mutually exclusive categories.
}
\setlength{\tabcolsep}{3pt}
\renewcommand{\arraystretch}{1.08}

\begin{tabular}{@{}lrrrrrrrrr@{}}
\toprule
& \multicolumn{3}{c}{Independent}
& \multicolumn{3}{c}{Retained revisions}
& \multicolumn{3}{c}{All-correct revisions} \\
\cmidrule(lr){2-4}
\cmidrule(lr){5-7}
\cmidrule(lr){8-10}
Dataset
& $N$ & No EOS & Unparsed
& $N$ & No EOS & Unparsed
& $N$ & No EOS & Unparsed \\
\midrule
MedQA
& 900 & 1 & 1
& 2{,}880 & 1 & 1
& 102 & 0 & 0 \\
ARC-C
& 3{,}495 & 2 & 2
& 2{,}352 & 9 & 9
& 832 & 0 & 0 \\
GSM8K
& 3{,}957 & 2 & 0
& 2{,}256 & 2 & 0
& 938 & 0 & 0 \\
GPQA-D
& 594 & 13 & 13
& 2{,}832 & 12 & 12
& 495 & 2 & 2 \\
HumanEval+
& 492 & 12 & 12
& 672 & 48 & 49
& 96 & 1 & 1 \\
\bottomrule
\end{tabular}

\label{tab:raw_integrity}
\end{table}

%% file: tables/intervals.tex
\begin{table}[t]
\centering
\small
\caption{
Retained-subset accuracy, CR, and PR (\%), with marginal
95\% percentile intervals from 2,000 question-cluster
bootstrap resamples.
Between-condition differences are assessed separately
using paired bootstrap intervals.
}
\setlength{\tabcolsep}{4pt}
\begin{tabular}{llrrr}
\toprule
Dataset & Condition & $\mathrm{Acc}_{\rm ret}$ [95\% CI] & CR [95\% CI] & PR [95\% CI] \\
\midrule
MedQA & No Message & 25.69 [20.69, 30.97] & 6.90 [2.14, 12.10] & 98.28 [95.54, 100.00] \\
 & Full Text & 27.22 [21.25, 33.34] & 80.17 [72.06, 88.14] & 42.24 [32.00, 52.90] \\
& StateBridge
& 27.78 [21.67, 34.17]
& 54.31 [42.62, 65.45]
& 71.55 [61.70, 81.03] \\
 & LatentMAS & 24.72 [19.72, 29.86] & 69.83 [60.20, 79.09] & 32.76 [24.53, 40.77] \\
\midrule
ARC-C & No Message & 27.89 [21.93, 34.01] & 8.16 [2.94, 14.82] & 92.86 [86.73, 97.87] \\
 & Full Text & 30.27 [23.64, 37.93] & 80.61 [71.15, 89.77] & 34.69 [24.00, 45.92] \\
 & StateBridge & 29.93 [23.30, 37.41] & 69.39 [59.00, 80.21] & 44.90 [33.33, 57.55] \\
 & LatentMAS & 28.23 [21.76, 35.20] & 58.16 [48.21, 67.59] & 45.92 [35.18, 57.32] \\
\midrule
GSM8K & No Message & 26.24 [19.68, 32.45] & 13.04 [4.65, 23.26] & 92.39 [85.11, 98.11] \\
 & Full Text & 25.35 [17.91, 32.45] & 51.09 [37.80, 64.45] & 52.17 [38.68, 65.72] \\
 & StateBridge & 26.95 [19.85, 33.87] & 39.13 [26.31, 52.17] & 73.91 [64.13, 83.33] \\
 & LatentMAS & 25.53 [18.79, 32.27] & 59.78 [46.94, 71.57] & 46.74 [34.21, 58.70] \\

 \midrule
GPQA-D & No Message
& 25.42 [20.20, 30.79]
& 14.04 [7.14, 22.03]
& 96.49 [92.73, 99.18] \\
& Full Text
& 27.26 [21.19, 33.33]
& 74.56 [65.22, 83.33]
& 47.37 [36.11, 59.17] \\
& StateBridge
& 26.84 [20.76, 32.77]
& 64.04 [54.10, 73.08]
& 51.75 [41.23, 63.08] \\
& LatentMAS
& 25.14 [19.49, 31.21]
& 61.40 [50.82, 71.56]
& 50.88 [40.35, 61.54] \\
\midrule
HumanEval+ & No Message & 42.26 [27.98, 55.95] & 57.14 [36.67, 76.67] & 92.86 [82.14, 100.00] \\
 & Full Text & 45.24 [29.17, 60.12] & 75.00 [53.57, 92.86] & 92.86 [81.82, 100.00] \\
 & StateBridge & 43.45 [28.57, 57.74] & 67.86 [46.87, 85.71] & 96.43 [87.50, 100.00] \\
 & LatentMAS & 39.88 [25.60, 53.57] & 85.71 [72.22, 96.43] & 60.71 [39.29, 81.82] \\
\bottomrule
\end{tabular}

\label{tab:intervals}
\end{table}

%% file: tables/paired.tex
\begin{table}[t]
\centering
\small
\caption{Matched outcomes against No Message over retained directions. Gain denotes No Message wrong/channel correct; loss denotes the reverse. These are answer-correctness transitions, not proof of sender-answer adoption.}
\setlength{\tabcolsep}{4pt}
\begin{tabular}{llrrrrrr}
\toprule
Dataset & Channel & $N$ & Both $\checkmark$ & Gain & Loss & Both $\times$ & Net \\
\midrule
MedQA & Full Text & 720 & 110 & 86 & 75 & 449 & +11 \\
 & StateBridge & 720 & 144 & 56 & 41 & 479 & +15 \\
 & LatentMAS & 720 & 98 & 80 & 87 & 455 & -7 \\
\midrule
ARC-C & Full Text & 588 & 102 & 76 & 62 & 348 & +14 \\
 & StateBridge & 588 & 112 & 64 & 52 & 360 & +12 \\
 & LatentMAS & 588 & 107 & 59 & 57 & 365 & +2 \\
\midrule
GSM8K & Full Text & 564 & 103 & 40 & 45 & 376 & -5 \\
 & StateBridge & 564 & 122 & 30 & 26 & 386 & +4 \\
 & LatentMAS & 564 & 94 & 50 & 54 & 366 & -4 \\
 \midrule
GPQA-D & Full Text
& 708 & 115 & 78 & 65 & 450 & +13 \\
& StateBridge
& 708 & 121 & 69 & 59 & 459 & +10 \\
& LatentMAS
& 708 & 116 & 62 & 64 & 466 & -2 \\
\midrule
HumanEval+ & Full Text & 168 & 63 & 13 & 8 & 84 & +5 \\
 & StateBridge & 168 & 62 & 11 & 9 & 86 & +2 \\
 & LatentMAS & 168 & 54 & 13 & 17 & 84 & -4 \\
\bottomrule
\end{tabular}

\label{tab:paired}
\end{table}

%% file: tables/transitions.tex
\begin{table}[t]
\centering
\small
\caption{
Answer-identity categories on retained directions.
Categories are assigned in the order Invalid, Prior,
Sender, and Third.
Sender therefore denotes a match that differs from the
receiver's initial answer.
Third also includes changed valid outputs when the sender's
initial answer is unparsed.
Program comparisons use string identity rather than
functional equivalence.
}
\setlength{\tabcolsep}{4pt}
\begin{tabular}{llrrrrr}
\toprule
Dataset & Condition & $N$ & Prior & Sender & Third & Invalid \\
\midrule
MedQA & No Message & 720 & 693 & 11 & 16 & 0 \\
 & Full Text & 720 & 496 & 220 & 3 & 1 \\
& StateBridge & 720 & 588 & 128 & 4 & 0 \\
 & LatentMAS & 720 & 503 & 207 & 10 & 0 \\
\midrule
ARC-C & No Message & 588 & 556 & 15 & 13 & 4 \\
 & Full Text & 588 & 429 & 157 & 1 & 1 \\
 & StateBridge & 588 & 455 & 129 & 2 & 2 \\
 & LatentMAS & 588 & 459 & 120 & 7 & 2 \\
\midrule
GSM8K & No Message & 564 & 524 & 20 & 20 & 0 \\
 & Full Text & 564 & 446 & 108 & 10 & 0 \\
 & StateBridge & 564 & 485 & 69 & 10 & 0 \\
 & LatentMAS & 564 & 437 & 118 & 9 & 0 \\

 \midrule
GPQA-D & No Message
& 708 & 637 & 28 & 41 & 2 \\
& Full Text
& 708 & 477 & 204 & 26 & 1 \\
& StateBridge
& 708 & 495 & 182 & 30 & 1 \\
& LatentMAS
& 708 & 511 & 170 & 19 & 8 \\
\midrule
HumanEval+ & No Message & 168 & 110 & 2 & 42 & 14 \\
 & Full Text & 168 & 97 & 19 & 42 & 10 \\
 & StateBridge & 168 & 105 & 10 & 41 & 12 \\
 & LatentMAS & 168 & 84 & 28 & 43 & 13 \\
\bottomrule
\end{tabular}

\label{tab:transitions}
\end{table}

%% file: tables/cost.tex
\begin{table}[t]
\centering
\small
\caption{Logged per-record means, not a controlled throughput comparison. Prompt-token accounting does not include all latent-prefix computation. Construction costs are not uniformly available. LatentMAS payload zeros in the source export are suppressed because they conflict with the documented KV-prefix transmission.}
\setlength{\tabcolsep}{4pt}
\begin{tabular}{llrrlr}
\toprule
Dataset & Condition & Prompt tok. & Output tok. & Message size & Revision (s) \\
\midrule
MedQA & No Message & 912.2 & 884.5 & -- & 19.93 \\
 & Full Text & 1438.2 & 978.7 & 518.1 tok. & 23.20 \\
 & StateBridge & 983.2 & 958.3 & 320.0 KiB & 21.73 \\
 & LatentMAS & 919.2 & 582.6 & Not recovered & 14.45 \\
\midrule
ARC-C & No Message & 709.7 & 807.5 & -- & 20.73 \\
 & Full Text & 1195.8 & 760.1 & 478.2 tok. & 16.40 \\
 & StateBridge & 780.7 & 773.6 & 320.0 KiB & 16.30 \\
 & LatentMAS & 716.7 & 570.0 & Not recovered & 14.69 \\
\midrule
GSM8K & No Message & 648.0 & 1012.5 & -- & 22.25 \\
 & Full Text & 1060.8 & 954.9 & 404.9 tok. & 20.81 \\
 & StateBridge & 719.0 & 967.3 & 320.0 KiB & 21.27 \\
 & LatentMAS & 655.0 & 502.3 & Not recovered & 12.95 \\
 \midrule
GPQA-D & No Message
& 1576 & 2909 & -- & 53.37 \\
& Full Text
& 2742 & 2216 & 1158.3 tok. & 41.62 \\
& StateBridge
& 1647 & 2539 & 319.8 KiB & 46.03 \\
& LatentMAS
& 1583 & 921 & Not recovered & 23.46 \\
\midrule
HumanEval+ & No Message & 2613.0 & 4422.4 & -- & 119.21 \\
 & Full Text & 4894.7 & 3878.4 & 2273.7 tok. & 106.57 \\
 & StateBridge & 2682.8 & 4102.9 & 314.0 KiB & 101.09 \\
 & LatentMAS & 2620.0 & 2345.6 & Not recovered & 86.10 \\
\bottomrule
\end{tabular}

\label{tab:cost}
\end{table}

%% file: tables/humaneval.tex
\begin{table}[t]
\centering
\caption{HumanEval+ supplementary results. Accuracy and SI are percentages; the remaining columns show success counts. All conditions evaluate the same 168 retained directions.}
\small
\setlength{\tabcolsep}{4pt}
\begin{tabular}{lrrrrrr}
\toprule
Condition & $\mathrm{Acc}_{\rm ret}$ & CR & PR & SI & SR & SCR \\
\midrule
No Message & 42.26 & 16/28 & 26/28 & 75.00 & 9/92 & 20/20 \\
Full Text & 45.24 & 21/28 & 26/28 & 83.93 & 9/92 & 20/20 \\
StateBridge & 43.45 & 19/28 & 27/28 & 82.14 & 8/92 & 19/20 \\
LatentMAS & 39.88 & 24/28 & 17/28 & 73.21 & 6/92 & 20/20 \\
\bottomrule
\end{tabular}

\label{tab:humaneval}
\end{table}

%% file: tables/extrapolation.tex
\begin{table}[t]
\centering
\small
\caption{Provisional full-set accuracy estimates, not measured full-set accuracy. Skipped-sample entries show correct/observed outcomes on all-correct triples. Coverage and estimates are percentages. The sampled directions need not constitute a probability sample of the unobserved directions. Exported interval bounds are not reproduced as confidence intervals; see text.}
\setlength{\tabcolsep}{4pt}
\begin{tabular}{llrrrrr}
\toprule
Dataset & Condition & Retained & Skipped sample & Imputed & Coverage & Estimate \\
\midrule
MedQA & No Message & 720 & 34/34 & 1046 & 41.89 & 70.28 \\
 & Full Text & 720 & 34/34 & 1046 & 41.89 & 70.89 \\
 & StateBridge & 720 & - & 1080 & 40.00 & 71.11 \\
 & LatentMAS & 720 & 34/34 & 1046 & 41.89 & 69.89 \\
\midrule
ARC-C & No Message & 588 & 208/208 & 6194 & 11.39 & 93.93 \\
 & Full Text & 588 & 208/208 & 6194 & 11.39 & 94.13 \\
 & StateBridge & 588 & 208/208 & 6194 & 11.39 & 94.11 \\
 & LatentMAS & 588 & 207/208 & 6194 & 11.39 & 93.52 \\
\midrule
GSM8K & No Message & 564 & 235/235 & 7115 & 10.10 & 94.74 \\
 & Full Text & 564 & 235/235 & 7115 & 10.10 & 94.68 \\
 & StateBridge & 564 & 235/235 & 7115 & 10.10 & 94.79 \\
 & LatentMAS & 564 & 233/233 & 7117 & 10.07 & 94.69 \\

 \midrule
GPQA-D & No Message
& 708 & 121/124 & 356 & 70.03 & 54.58 \\
& Full Text
& 708 & 123/124 & 356 & 70.03 & 56.32 \\
& StateBridge
& 708 & 122/124 & 356 & 70.03 & 55.75 \\
& LatentMAS
& 708 & 122/123 & 357 & 69.95 & 55.06 \\
\midrule
HumanEval+ & No Message & 168 & 23/24 & 792 & 19.51 & 86.69 \\
 & Full Text & 168 & 24/24 & 792 & 19.51 & 90.65 \\
 & StateBridge & 168 & 24/24 & 792 & 19.51 & 90.35 \\
 & LatentMAS & 168 & 24/24 & 792 & 19.51 & 89.74 \\
\bottomrule
\end{tabular}

\label{tab:extrapolation}
\end{table}

%% file: tables/directions.tex
\begingroup
\small
\setlength{\tabcolsep}{4pt}
\begin{longtable}{lllrrrrr}
\caption{Per-direction outcomes. Conditional columns show successes/denominator.}\label{tab:directions}\\
\toprule Dataset & Channel & Direction & $N$ & CR & PR & SR & SCR \\
\midrule\endfirsthead
\multicolumn{8}{l}{\small Table \thetable\ continued}\\
\toprule Dataset & Channel & Direction & $N$ & CR & PR & SR & SCR \\
\midrule\endhead
\midrule\multicolumn{8}{r}{Continued on next page}\\\endfoot
\bottomrule\endlastfoot
MedQA & No Message & A$\to$B & 120 & 1/18 & 21/22 & 2/72 & 8/8 \\
 &  & A$\to$C & 120 & 1/18 & 19/20 & 2/74 & 8/8 \\
 &  & B$\to$A & 120 & 0/22 & 18/18 & 3/72 & 8/8 \\
 &  & B$\to$C & 120 & 1/20 & 18/18 & 3/72 & 10/10 \\
 &  & C$\to$A & 120 & 2/20 & 18/18 & 1/74 & 8/8 \\
 &  & C$\to$B & 120 & 3/18 & 20/20 & 0/72 & 10/10 \\
\addlinespace[2pt]
MedQA & Full Text & A$\to$B & 120 & 12/18 & 8/22 & 1/72 & 8/8 \\
 &  & A$\to$C & 120 & 16/18 & 9/20 & 0/74 & 8/8 \\
 &  & B$\to$A & 120 & 18/22 & 5/18 & 0/72 & 8/8 \\
 &  & B$\to$C & 120 & 14/20 & 11/18 & 0/72 & 10/10 \\
 &  & C$\to$A & 120 & 16/20 & 9/18 & 1/74 & 8/8 \\
 &  & C$\to$B & 120 & 17/18 & 7/20 & 0/72 & 10/10 \\
\addlinespace[2pt]
MedQA & StateBridge
& A$\to$B & 120 & 9/18 & 14/22 & 1/72 & 8/8 \\
& & A$\to$C
& 120 & 12/18 & 17/20 & 1/74 & 8/8 \\
& & B$\to$A
& 120 & 11/22 & 12/18 & 0/72 & 8/8 \\
& & B$\to$C
& 120 & 10/20 & 15/18 & 0/72 & 10/10 \\
& & C$\to$A
& 120 & 11/20 & 12/18 & 0/74 & 8/8 \\
& & C$\to$B
& 120 & 10/18 & 13/20 & 0/72 & 10/10 \\
\addlinespace[2pt]
MedQA & LatentMAS & A$\to$B & 120 & 14/18 & 7/22 & 1/72 & 8/8 \\
 &  & A$\to$C & 120 & 10/18 & 7/20 & 1/74 & 8/8 \\
 &  & B$\to$A & 120 & 16/22 & 4/18 & 1/72 & 8/8 \\
 &  & B$\to$C & 120 & 13/20 & 8/18 & 1/72 & 10/10 \\
 &  & C$\to$A & 120 & 14/20 & 5/18 & 3/74 & 8/8 \\
 &  & C$\to$B & 120 & 14/18 & 7/20 & 0/72 & 10/10 \\
\addlinespace[2pt]
ARC-C & No Message & A$\to$B & 98 & 3/16 & 14/15 & 1/57 & 10/10 \\
 &  & A$\to$C & 98 & 0/15 & 16/19 & 1/53 & 11/11 \\
 &  & B$\to$A & 98 & 0/15 & 16/16 & 0/57 & 9/10 \\
 &  & B$\to$C & 98 & 3/14 & 19/19 & 1/54 & 10/11 \\
 &  & C$\to$A & 98 & 1/19 & 13/15 & 0/53 & 10/11 \\
 &  & C$\to$B & 98 & 1/19 & 13/14 & 2/54 & 10/11 \\
\addlinespace[2pt]
ARC-C & Full Text & A$\to$B & 98 & 13/16 & 2/15 & 1/57 & 10/10 \\
 &  & A$\to$C & 98 & 13/15 & 3/19 & 0/53 & 11/11 \\
 &  & B$\to$A & 98 & 12/15 & 7/16 & 0/57 & 10/10 \\
 &  & B$\to$C & 98 & 10/14 & 6/19 & 0/54 & 11/11 \\
 &  & C$\to$A & 98 & 15/19 & 8/15 & 0/53 & 11/11 \\
 &  & C$\to$B & 98 & 16/19 & 8/14 & 0/54 & 11/11 \\
\addlinespace[2pt]
ARC-C & StateBridge & A$\to$B & 98 & 8/16 & 5/15 & 0/57 & 10/10 \\
 &  & A$\to$C & 98 & 13/15 & 8/19 & 1/53 & 11/11 \\
 &  & B$\to$A & 98 & 11/15 & 8/16 & 0/57 & 10/10 \\
 &  & B$\to$C & 98 & 11/14 & 8/19 & 0/54 & 11/11 \\
 &  & C$\to$A & 98 & 11/19 & 7/15 & 0/53 & 11/11 \\
 &  & C$\to$B & 98 & 14/19 & 8/14 & 0/54 & 10/11 \\
\addlinespace[2pt]
ARC-C & LatentMAS & A$\to$B & 98 & 8/16 & 6/15 & 0/57 & 10/10 \\
 &  & A$\to$C & 98 & 9/15 & 6/19 & 0/53 & 11/11 \\
 &  & B$\to$A & 98 & 9/15 & 7/16 & 2/57 & 9/10 \\
 &  & B$\to$C & 98 & 8/14 & 10/19 & 1/54 & 10/11 \\
 &  & C$\to$A & 98 & 11/19 & 8/15 & 0/53 & 10/11 \\
 &  & C$\to$B & 98 & 12/19 & 8/14 & 0/54 & 11/11 \\
\addlinespace[2pt]
GSM8K & No Message & A$\to$B & 94 & 1/16 & 13/15 & 0/56 & 7/7 \\
 &  & A$\to$C & 94 & 4/15 & 13/16 & 1/55 & 8/8 \\
 &  & B$\to$A & 94 & 2/15 & 16/16 & 1/56 & 6/7 \\
 &  & B$\to$C & 94 & 3/14 & 14/16 & 2/56 & 8/8 \\
 &  & C$\to$A & 94 & 1/16 & 15/15 & 2/55 & 8/8 \\
 &  & C$\to$B & 94 & 1/16 & 14/14 & 0/56 & 8/8 \\
\addlinespace[2pt]
GSM8K & Full Text & A$\to$B & 94 & 5/16 & 7/15 & 0/56 & 7/7 \\
 &  & A$\to$C & 94 & 9/15 & 7/16 & 0/55 & 8/8 \\
 &  & B$\to$A & 94 & 6/15 & 7/16 & 0/56 & 7/7 \\
 &  & B$\to$C & 94 & 8/14 & 12/16 & 2/56 & 8/8 \\
 &  & C$\to$A & 94 & 9/16 & 7/15 & 0/55 & 8/8 \\
 &  & C$\to$B & 94 & 10/16 & 8/14 & 0/56 & 8/8 \\
\addlinespace[2pt]
GSM8K & StateBridge & A$\to$B & 94 & 6/16 & 10/15 & 0/56 & 6/7 \\
 &  & A$\to$C & 94 & 7/15 & 13/16 & 2/55 & 8/8 \\
 &  & B$\to$A & 94 & 4/15 & 12/16 & 1/56 & 7/7 \\
 &  & B$\to$C & 94 & 8/14 & 12/16 & 0/56 & 8/8 \\
 &  & C$\to$A & 94 & 5/16 & 9/15 & 0/55 & 8/8 \\
 &  & C$\to$B & 94 & 6/16 & 12/14 & 0/56 & 8/8 \\
\addlinespace[2pt]
GSM8K & LatentMAS & A$\to$B & 94 & 10/16 & 7/15 & 1/56 & 6/7 \\
 &  & A$\to$C & 94 & 9/15 & 9/16 & 0/55 & 8/8 \\
 &  & B$\to$A & 94 & 9/15 & 7/16 & 1/56 & 6/7 \\
 &  & B$\to$C & 94 & 9/14 & 7/16 & 0/56 & 8/8 \\
 &  & C$\to$A & 94 & 8/16 & 7/15 & 0/55 & 8/8 \\
 &  & C$\to$B & 94 & 10/16 & 6/14 & 0/56 & 8/8 \\
\addlinespace[2pt]
GPQA-D & No Message & A$\to$B & 118 & 3/19 & 23/23 & 1/70 & 5/6 \\
 &  & A$\to$C & 118 & 1/15 & 16/17 & 3/76 & 9/10 \\
 &  & B$\to$A & 118 & 5/23 & 19/19 & 1/70 & 6/6 \\
 &  & B$\to$C & 118 & 3/21 & 16/19 & 1/70 & 8/8 \\
 &  & C$\to$A & 118 & 3/17 & 15/15 & 3/76 & 9/10 \\
 &  & C$\to$B & 118 & 1/19 & 21/21 & 1/70 & 7/8 \\
\addlinespace[2pt]
GPQA-D & Full Text & A$\to$B & 118 & 13/19 & 12/23 & 0/70 & 6/6 \\
 &  & A$\to$C & 118 & 11/15 & 9/17 & 3/76 & 10/10 \\
 &  & B$\to$A & 118 & 15/23 & 7/19 & 0/70 & 6/6 \\
 &  & B$\to$C & 118 & 18/21 & 7/19 & 1/70 & 8/8 \\
 &  & C$\to$A & 118 & 13/17 & 6/15 & 1/76 & 10/10 \\
 &  & C$\to$B & 118 & 15/19 & 13/21 & 1/70 & 8/8 \\
\addlinespace[2pt]
GPQA-D & StateBridge & A$\to$B & 118 & 12/19 & 10/23 & 3/70 & 6/6 \\
 &  & A$\to$C & 118 & 8/15 & 12/17 & 4/76 & 9/10 \\
 &  & B$\to$A & 118 & 19/23 & 8/19 & 0/70 & 6/6 \\
 &  & B$\to$C & 118 & 13/21 & 7/19 & 1/70 & 8/8 \\
 &  & C$\to$A & 118 & 8/17 & 10/15 & 2/76 & 10/10 \\
 &  & C$\to$B & 118 & 13/19 & 12/21 & 1/70 & 8/8 \\
\addlinespace[2pt]
GPQA-D & LatentMAS & A$\to$B & 118 & 12/19 & 14/23 & 0/70 & 5/6 \\
 &  & A$\to$C & 118 & 8/15 & 9/17 & 1/76 & 10/10 \\
 &  & B$\to$A & 118 & 15/23 & 5/19 & 1/70 & 6/6 \\
 &  & B$\to$C & 118 & 13/21 & 12/19 & 0/70 & 8/8 \\
 &  & C$\to$A & 118 & 10/17 & 7/15 & 2/76 & 9/10 \\
 &  & C$\to$B & 118 & 12/19 & 11/21 & 0/70 & 8/8 \\
\addlinespace[2pt]

HumanEval+ & No Message & A$\to$B & 28 & 3/5 & 3/3 & 2/15 & 5/5 \\
 &  & A$\to$C & 28 & 4/7 & 3/3 & 1/15 & 3/3 \\
 &  & B$\to$A & 28 & 1/3 & 5/5 & 1/15 & 5/5 \\
 &  & B$\to$C & 28 & 3/6 & 3/4 & 3/16 & 2/2 \\
 &  & C$\to$A & 28 & 2/3 & 6/7 & 2/15 & 3/3 \\
 &  & C$\to$B & 28 & 3/4 & 6/6 & 0/16 & 2/2 \\
\addlinespace[2pt]
HumanEval+ & Full Text & A$\to$B & 28 & 3/5 & 3/3 & 1/15 & 5/5 \\
 &  & A$\to$C & 28 & 6/7 & 3/3 & 1/15 & 3/3 \\
 &  & B$\to$A & 28 & 2/3 & 5/5 & 1/15 & 5/5 \\
 &  & B$\to$C & 28 & 4/6 & 4/4 & 3/16 & 2/2 \\
 &  & C$\to$A & 28 & 2/3 & 6/7 & 1/15 & 3/3 \\
 &  & C$\to$B & 28 & 4/4 & 5/6 & 2/16 & 2/2 \\
\addlinespace[2pt]
HumanEval+ & StateBridge & A$\to$B & 28 & 3/5 & 3/3 & 0/15 & 5/5 \\
 &  & A$\to$C & 28 & 4/7 & 2/3 & 2/15 & 2/3 \\
 &  & B$\to$A & 28 & 2/3 & 5/5 & 1/15 & 5/5 \\
 &  & B$\to$C & 28 & 4/6 & 4/4 & 3/16 & 2/2 \\
 &  & C$\to$A & 28 & 3/3 & 7/7 & 0/15 & 3/3 \\
 &  & C$\to$B & 28 & 3/4 & 6/6 & 2/16 & 2/2 \\
\addlinespace[2pt]
HumanEval+ & LatentMAS & A$\to$B & 28 & 4/5 & 2/3 & 1/15 & 5/5 \\
 &  & A$\to$C & 28 & 7/7 & 1/3 & 0/15 & 3/3 \\
 &  & B$\to$A & 28 & 2/3 & 3/5 & 0/15 & 5/5 \\
 &  & B$\to$C & 28 & 5/6 & 4/4 & 1/16 & 2/2 \\
 &  & C$\to$A & 28 & 3/3 & 4/7 & 2/15 & 3/3 \\
 &  & C$\to$B & 28 & 3/4 & 3/6 & 2/16 & 2/2 \\
\addlinespace[2pt]
\end{longtable}
\endgroup